\documentclass[journal]{IEEEtran} 

\usepackage{graphicx}
\graphicspath{ {Images/} }
\usepackage{xcolor}
\usepackage[caption=false]{subfig} 
\usepackage{float}
\usepackage{enumitem}
\usepackage{dblfloatfix} 

\usepackage{algorithm}
\usepackage[]{algpseudocode}
\usepackage{amsmath} 
\usepackage{amssymb}
\usepackage{latexsym}

\usepackage[colorlinks=true, citecolor=blue, urlcolor=black, linkcolor=magenta]{hyperref}
\usepackage{seqsplit}
\usepackage{xurl}
\usepackage{bbding} 
\usepackage{comment} 
\usepackage{soul}
\usepackage{tikz}
\usepackage{bbding} 

\usepackage{tabularray}
\usepackage{tabularx}
\usepackage{multirow}
\usepackage{rotating}
\usepackage{array} 
\newcolumntype{C}{>{\centering\arraybackslash}X}
\newcolumntype{L}{>{\raggedright\arraybackslash}X}
\newcolumntype{R}{>{\raggedright\arraybackslash}X}

\definecolor{palesilver}{rgb}{0.96, 0.96, 0.96}
\definecolor{pastelgray}{rgb}{0.81, 0.81, 0.77}
\definecolor{arsenic}{rgb}{0.23, 0.27, 0.29}

\usepackage{flushend}
\renewcommand\thesection{\arabic{section}}
\renewcommand\thesubsection{\thesection.\arabic{subsection}}
\renewcommand\thesubsubsection{\thesubsection.\arabic{subsubsection}}

\renewcommand\thesectiondis{\arabic{section}}
\renewcommand\thesubsectiondis{\thesectiondis.\arabic{subsection}}
\renewcommand\thesubsubsectiondis{\thesubsectiondis.\arabic{subsubsection}}

\newcommand{\revisedSG}[1]{{\color{black}{#1}}}

\definecolor{yellowcolor}{rgb}{1, 0.85, 0.4}
\definecolor{greencolor}{rgb}{0.58, .77, 0.49}
\definecolor{bluecolor}{rgb}{0.43, 0.65, 0.86}

\newcommand\YC[1]{%
  \tikz[baseline=(X.base)] 
    \node(X)[draw, shape=circle, inner sep=0.2, fill=yellowcolor, text=black] {\scriptsize #1};%
}
\newcommand\GC[1]{%
  \tikz[baseline=(X.base)] 
    \node(X)[draw, shape=circle, inner sep=0.2, fill=greencolor, text=black] {\scriptsize #1};%
}

\usepackage{microtype} 
\begin{document}
\microtypesetup{activate=true}

\title{\huge{Robust Vision Systems for Connected and Autonomous Vehicles: Security Challenges and Attack Vectors}}
\author{\IEEEauthorblockN{Sandeep Gupta\IEEEauthorrefmark{1} and
Roberto Passerone\IEEEauthorrefmark{2}}\\
\IEEEauthorblockA{\IEEEauthorrefmark{1}Centre for Secure Information Technologies (CSIT), Queen's University Belfast, UK}\\
\IEEEauthorblockA{\IEEEauthorrefmark{2}Department of Information Engineering and Computer Science, University of Trento, Italy}
}

\maketitle
\thispagestyle{plain}
\pagestyle{plain}

\begin{abstract}
This paper investigates the robustness of vision systems in Connected and Autonomous Vehicles (CAVs), which is critical for achieving Level‑5 autonomous driving. Safe and reliable CAV navigation depends on robust vision systems that enable accurate detection of objects, lane markings, and traffic signs. \revisedSG{This survey presents a reference architecture for CAV vision systems (CAVVS) and uses it to derive a system‑level threat model that maps assets, vulnerabilities, and attack points to three concrete attack surfaces (data, model, input) across the perception lifecycle. We examine attack vectors targeting each surface, rigorously evaluating their implications for confidentiality, integrity, and availability (CIA) and comparing them by adversary capability and practical exposure. We further outline atomic road events that can impact CAVVS robustness and propose a benchmark‑oriented framework of datasets, metrics, and evaluation protocols for standardized future assessment. Together, these contributions provide a structured basis for vulnerability assessment, principled defense design, and reproducible robustness evaluation in real‑world CAV vision systems.}
\end{abstract}


\section{Introduction}
Connected and Autonomous Vehicles (CAVs) are set to redefine mobility by enhancing accessibility, convenience, efficiency, and safety through full driving automation~\cite{chen2022milestones, sun2021survey}. According to the SAE J3016 standards, automation levels range from 0 (no automation) to 5 (full automation)~\cite{sae2014taxonomy}. Automotive industries and technology firms are heavily investing in research and development of CAVs to reach Level-5 autonomy. The standard defines Level-5 as a fully autonomous driving system capable of performing dynamic driving tasks under all roadway and environmental conditions for which the system is designed, without the need for human intervention. Level-5 autonomy requires a reliable and secure vision system that can perceive both static and dynamic objects along their path serving as the eyes and ears of CAVs~\cite{wang2023does, cao2021invisible}.

In CAVs, vision systems are typically equipped with sensors such as cameras, LiDAR, RADAR, and ultrasonic sensors~\cite{thakur2024depth, mahima2024toward}. These sensors detect and interpret information about nearby objects, road conditions, traffic signs, pedestrians, and other vehicles, thereby generating a comprehensive view of the surroundings. Furthermore, CAVs may encounter road incidents or situations that are spontaneous, discrete, and self-contained, collectively referred to as \textit{atomic road events}. Such incidents or situations can arise due to the approach of an emergency vehicle, pedestrian crossing or jaywalking, vehicle turning or lane changing, unplanned construction zone, vehicle breakdown or accident, and natural disaster or unexpected weather change. Consequently, a CAV requires a robust vision system for instantaneous situation understanding and response along with an alert broadcast to Vehicle-to-Everything (V2X) that can ensure safety and prevent a disaster.

To attain Level-5 autonomy, a vision system must seamlessly integrate visual perception with cognitive processes, facilitating precise environmental interpretation, informed decision-making, and reliable autonomous operation without human intervention~\cite{kemsaram2021model,long2022survey}. However, studies~\cite{he2019cooperative, marco2020multi} have highlighted numerous overarching challenges, including the lack of robustness and the black-box nature of deep learning models employed in CAV vision systems. For instance, there was a fatal incident in Florida involving a Tesla autonomous vehicle operating in Autopilot mode~\cite{girdhar2023cybersecurity}. Similarly, a self-driving Uber vehicle collided with a pedestrian in Tempe, Arizona. Pedestrian and cyclist detection, in particular, remains a significant challenge, as evidenced by recent KITTI benchmark results~\cite{he2019cooperative,liu2020computing}. Thus, a robust vision system is indispensable for CAVs to accurately perceive objects, lanes, and traffic signs, enabling safe and accident-free navigation~\cite{sharma2022cybersecurity}.

In this paper, we investigate the robustness of CAV vision systems (CAVVS), identifying the attack vectors that can compromise confidentiality, integrity, and availability (CIA) of CAVVS. We analyze the key sensors and vision components essential for CAV navigation and present a reference architecture for CAVVS (refer to Figure~\ref{fig:CAVVisionArch}). This architecture can facilitate the identification of attack surfaces, a thorough  understanding of attack vector dynamics and the systematic articulation of robust security measures to support compliance with the CIA triad in CAVVS.

\subsection{Motivation}
\noindent
This survey aims to answer and is guided by the following research questions:
\begin{enumerate}[leftmargin=*,label=\textbf{RQ\arabic*.}]
    \item What are the \emph{challenges} for a robust vision system design?
    \item What are the \emph{critical assets} in CAVVS, and what vulnerabilities could expose them to potential attacks?
    \item What \emph{commonalities} can be used to derive the attack surfaces?
    \item What \emph{strategies and mechanisms} can be leveraged to exploit the identified attack surfaces?
    \item What are the \emph{characteristics} of identified attack vectors?
\end{enumerate}

\revisedSG{
\noindent Table~\ref{tab:RQMap} maps each research question to the section in which it is addressed. The manuscript progresses from discussing the challenges for a robust vision system design (RQ1, Section~\ref{sec:CAVVSChallenges}), to identifying attack surfaces (RQ2--RQ3, Section~\ref{sec:TMASD}), and finally to analyzing the attack vectors that exploit them (RQ4--RQ5, Section~\ref{sec:AV}).
\begin{table}[!ht]
    \centering
    \scriptsize
    \caption{Mapping of research questions to manuscript sections}\label{tab:RQMap}
    \begin{tabular}{p{.08\linewidth} p{.67\linewidth} p{.1\linewidth}}\hline
    \textbf{RQ} & \textbf{Focus} & \textbf{Section}\\\hline
    RQ1 & Challenges for robust vision system design & \ref{sec:CAVVSChallenges}\\\hline
    RQ2 & Critical assets and vulnerabilities & \ref{sec:TMASD}\\\hline
    RQ3 & Commonalities used to derive attack surfaces & \ref{sec:TMASD}\\\hline
    RQ4 & Strategies and mechanisms exploiting attack surfaces & \ref{sec:AV}\\\hline
    RQ5 & Characteristics of attack vectors & \ref{sec:AV}\\\hline
    \end{tabular}
\end{table}
}

\subsection{Contributions}
\revisedSG{
\noindent The main contributions of the paper are outlined as follows: 
\begin{itemize}[leftmargin=*]
    \item Discuss challenges in object detection, classification, and prediction for CAV vision systems, situating recent evidence on physical adversarial patches, camera-LiDAR fusion spoofing, and point-cloud/BEV adversarial attacks within the CAVVS context (Section~\ref{sec:CAVVSChallenges}).
    \item Present a reference architecture for CAVVS, delineating the responsibilities of the \textit{perception}, \textit{data processing and computation}, and \textit{motion estimation} layers, and compare CAVVS sensor modalities and industrial sensor suites in terms of outputs, limitations, and the access and capability required to exploit them (Sections~\ref{sec:CAVVS}--\ref{sec:SystemArch}).
    \item Define a threat model that identifies the critical assets and vulnerabilities of CAVVS and derives its data, model, and input attack surfaces by mapping known attack points from a baseline sensor-based recognition system (Section~\ref{sec:TMASD}).
    \item Derive and analyze eight attack vectors, i.e., \textit{data poisoning}, \textit{data exfiltration}, \textit{model extraction}, \textit{logic corruption}, \textit{membership inference}, \textit{side-channel attacks}, \textit{evasion}, and \textit{man-in-the-middle} (MitM) attacks, each characterized by mechanism/strategy, targeted architectural layer(s), and CIA impact, and further compared across adversary capability (Section~\ref{sec:AV}).
    \item Outline atomic road events that challenge robust CAVVS design and propose a benchmark-oriented framework, namely, candidate datasets, metrics, and evaluation protocols, for standardized robustness assessment of CAVVS as a direction for future research (Section~\ref{sec:OCRD}).
\end{itemize}
}

\revisedSG{
\begin{table*}[!ht]
    \centering
    \hyphenpenalty 10000
    \scriptsize
    \caption{Comparison of recent CAVs security surveys based on research topic, attacks/attack Surfaces, vision system, ML-security, and CIA properties of vision system}\label{tab:RecentSurveys}
    \begin{tabularx}{1\linewidth}{p{.12\linewidth} p{.02\linewidth} p{.31\linewidth} p{.3\linewidth} p{.025\linewidth} p{.025\linewidth} p{.025\linewidth}}\hline
    \textbf{Reference} & \textbf{Year} & \textbf{Research topic covered} & \textbf{Attacks/Attack Surfaces} & \rotatebox[origin=c]{60}{\parbox[c]{1cm}{\textbf{Vision System}}} & \rotatebox[origin=c]{60}{\parbox[c]{1cm}{\textbf{ML-Security}}} & \rotatebox[origin=c]{60}{\parbox[c]{.8cm}{\textbf{CIA Triad}}}\\\hline
    \textsc{This survey} & - & Investigate the robustness of CAV vision systems (CAVVS) & Vision system life cycle (Refer to Figure~\ref{fig:CAVVSLifecycle}) & \Checkmark & \Checkmark & \Checkmark\\\hline
    Gu et al.~\cite{gu2025cyberattack} & 2025 & Propose a generative-based human-in-the-loop trajectory prediction approach for connected vehicle safety under spoofing cyberattacks. & Spoofing cyberattacks on connected vehicles; trajectory prediction under attack. & \Checkmark & - & \Checkmark \\\hline
    Chib and Singh~\cite{chib2023recent} & 2024 & End-to-End autonomous driving stack involving deep neural network (DNN) for driving process from perception to control. & Adversarial attacks on sensors. & - & - & -\\\hline
    Ellis et al.~\cite{ellis2024machine} & 2024 & ML security challenges from the perspective of a system of interconnected vehicles. & Attacks scenarios originating from both adversarial behaviour and conditions. & \Checkmark & -& - \\\hline
    Wang et al.~\cite{wang2024data} & 2024 & Examine data poisoning attack models in intelligent transportation systems (ITS) according to the data source they are targeting and the functions the data supports & Data poisoning attacks against vehicular data, i.e., localization, perception, planning and control. & \Checkmark & - & - \\\hline
    Chen et al.~\cite{chen2022milestones} & 2023 & Survey of surveys of the development of autonomous driving & - & - & - & -\\\hline
    Girdhar et al.~\cite{girdhar2023cybersecurity} & 2023 & Outline vulnerabilities and limitations of autonomous vehicle. & Artificial intelligence usage in autonomous driving. & \Checkmark & \Checkmark & - \\\hline
    Abdo et al.~\cite{abdo2023cybersecurity} & 2023 & Cyber attacks targeting CAVs through sensors and peripherals such as cameras, radar, LiDAR, and GPS. & Multi-vehicle and transportation system. & \Checkmark & - & - \\\hline
    Sharma and Gillanders~\cite{sharma2022cybersecurity} & 2022 & Examine safety and forensics standards in intelligent transportation system. & Cybersecurity and forensics challenges. & \Checkmark & - & \Checkmark \\\hline
    Deng et al.~\cite{deng2021deep} & 2021 & Investigate attacks on the pipeline of deep learning-based autonomous driving systems. & Physical and adversarial attacks. & \Checkmark & \Checkmark & - \\\hline
    Sun et al.~\cite{sun2021survey} & 2021 & Investigate cyber-security in the environment of CAVs highlighting security problems and challenges. & In-vehicle network attacks, vehicle to everything network attacks & \Checkmark & - & - \\\hline
    Pham and Xiong~\cite{pham2021survey} & 2021 & Discuss attack models and defense strategies for attack sequences on the CAVs components & CAV components vulnerability and their exploitation by adversaries. & \Checkmark & - & \Checkmark \\\hline
    Cao et al.~\cite{cao2021invisible} & 2021 & Investigate security of multi-sensor fusion (MSF) based perception under physical-world adversarial attacks targeting both camera and LiDAR. & Physical adversarial attacks on MSF perception, including 3D-printed adversarial objects and laser spoofing. & \Checkmark & \Checkmark & \Checkmark \\\hline
    Jiang et al.~\cite{jiang2020poisoning} & 2020 & Attacks against deep learning algorithms in traffic sign recognition system. & Poisoning and evasion attack with particle swarm optimization & - & - & \Checkmark\\\hline
    \end{tabularx}
\end{table*}

\subsection{Positioning of this work}
In comparison to previous surveys related to CAVs listed in Table~\ref{tab:RecentSurveys}, we focus specifically on the security aspects of the vision system in CAVs, linking the functional architecture, attack surfaces, attack vectors, and security consequences within a unified framework, which is directly useful for security engineers and system architects. Two surveys most closely align with our scope. Sharma and Gillanders~\cite{sharma2022cybersecurity} examine CIA properties from the perspective of forensics and safety standards, treating CIA primarily as a static compliance classification rather than relating it to the mechanics of the vision processing pipeline. Similarly, Pham and Xiong~\cite{pham2021survey} discuss CIA-relevant defense strategies at the level of CAV components in general, without distinguishing attack surfaces across a vision-specific life cycle. Notably, Cao et al.~\cite{cao2021invisible} demonstrate physical adversarial attacks on multi-sensor fusion perception, but their scope is limited to a specific attack class rather than a comprehensive lifecycle-based taxonomy with CIA analysis. Meanwhile, surveys that engage substantively with ML security, including Girdhar et al.~\cite{girdhar2023cybersecurity} and Deng et al.~\cite{deng2021deep}, do not examine the CIA implications of the attacks discussed. This survey differs from existing reviews in three respects. First, rather than treating adversarial machine learning as an isolated model-level problem, we examine attacks in relation to the complete CAVVS processing pipeline. Second, rather than surveying cybersecurity across the entire autonomous-driving stack, we restrict our focus on the vision-related components and interfaces that influence environmental perception and motion estimation. Third, rather than organizing attacks only by technique or sensing modality, we relate each attack vector to its attack surface, adversary capability, affected architectural component, and CIA consequence.
}

\revisedSG{
\section{Challenges in object detection, classification, and prediction}\label{sec:CAVVSChallenges}
CAVs rely on deep-learning and traditional machine-learning models to process multimodal sensor data, enabling robust environmental perception, accurate scene interpretation, and real-time navigation. However, studies~\cite{he2019cooperative, ellis2024machine, zhang2023atta} have highlighted limitations in the ability of these models to perform critical tasks, including object detection, traffic sign classification, and steering angle prediction, across both laboratory and real-world settings. In particular, object detection models evaluated on the KITTI autonomous driving benchmark have demonstrated significant challenges in accurately detecting and tracking objects under complex real-world conditions, including adverse weather, occlusion, and variations in object scale. Since accurate and real-time object detection is fundamental to the safe deployment of autonomous driving systems across diverse environmental conditions, vulnerabilities in the detection of critical objects, such as vehicles, lanes, and pedestrians, can have potentially devastating consequences~\cite{shi2021computing}.

Wang et al.~\cite{wang2023does} report that physical adversarial object-evasion attacks, particularly those involving malicious patches, can cause CAVVS to misclassify or overlook road objects during detection. Building on this line of work, Zhu et al.~\cite{zhu2022infrared} develop an adversarial ``QR code'' pattern that successfully deceives infrared pedestrian detectors across a range of viewing angles in both digital and physical settings. As shown in Figure~\ref{fig:DetPersonInfra}, the YOLOv3 detector fails to detect pedestrians when the adversarial patch is applied. Going beyond patch-based attacks, Wei et al.~\cite{wei2023unified} propose a boundary-limited shape optimisation method for generating compact and smooth adversarial shapes that enable cross-modal physical attacks. Figure~\ref{fig:DetPersonUnified} illustrates a cross-modal adversarial patch that simultaneously evades visible and infrared pedestrian detectors in the physical world.

\begin{figure}[!ht]
    \captionsetup{format=hang,font=small, margin=5pt}
    \centering
    \subfloat[YOLOv3 detector fails to detect individuals with the patch applied~\cite{zhu2022infrared}\label{fig:DetPersonInfra}]{
        \includegraphics[width=0.46\linewidth, keepaspectratio]{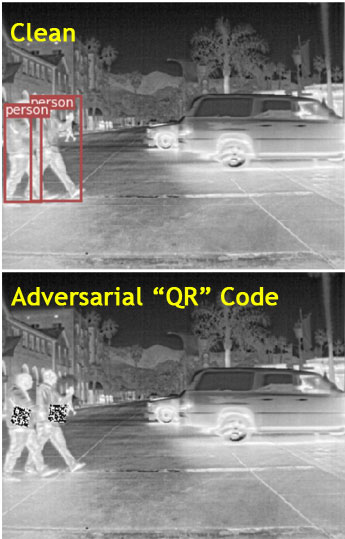}
    }
    \subfloat[Detectors like YOLOv3 and Faster RCNN unified deceived by cross-modal adversarial patches~\cite{wei2023unified}\label{fig:DetPersonUnified}]{
        \includegraphics[width=0.46\linewidth, keepaspectratio]{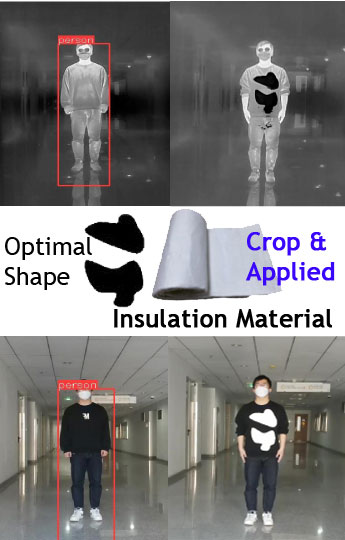}
    }
    \caption{Illustration of patch-based attacks on object detection models}\label{fig:CAVVSDetection1}
\end{figure}

Chen et al.~\cite{chen2019shapeshifter} proposed ShapeShifter, an adversarial attack using iterative change-of-variable and Expectation over Transformation. The Faster R-CNN based detector consistently detected adversarially perturbed stop signs generated using ShapeShifter as objects other than stop sign under physical conditions.
Hallyburton et al.~\cite{hallyburton2022security} introduce the frustum attack, which compromises camera-LiDAR fusion by preserving semantic consistency between camera and LiDAR data to spoof the sensor fusion process (refer to Figure~\ref{fig:Frustum}) for launching evasion attack on CAVVS. The authors evaluate the frustum attack on three distinct LiDAR-only architectures and five models spanning three different camera-LiDAR fusion architectures, including fusion at the semantic, feature, and tracking levels. Lin et al.~\cite{lin2024phade} describe phantom spoofing attacks, in which adversaries use electronic displays to create deceptive objects or signs such as fake pedestrians, vehicles, or incorrect speed limits (refer to Figure~\ref{fig:Phantom}) causing CAVs to apply an emergency brake or to exceed the speed. Yang et al.~\cite{yang2021robust} conduct an experiment to compromise LiDAR perception by placing 3D-printed objects, such as polyhedrons, which were misclassified as vehicles by the detection models.
\begin{figure}[!ht]
    \captionsetup{format=hang,font=small, margin=5pt}
    \centering
    \subfloat[Apollo perception system spoofing outcomes. Two LiDAR snapshots of detected objects before and after the frustum attack~\cite{hallyburton2022security}\label{fig:Frustum}]{
        \includegraphics[width=0.8\linewidth, keepaspectratio]{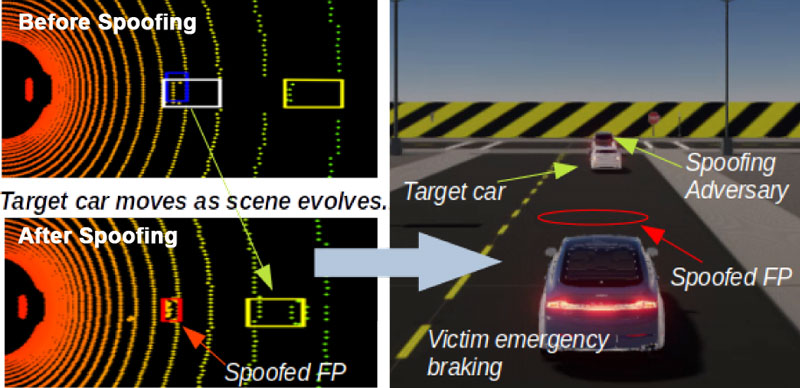}
    }\\
    \subfloat[Phantom spoofing attacks using deceptive objects or signs~\cite{lin2024phade}\label{fig:Phantom}]{
        \includegraphics[width=0.8\linewidth, keepaspectratio]{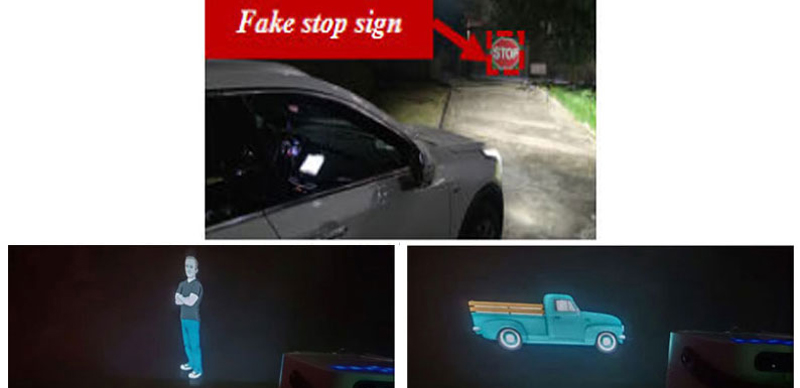}
    }
    \caption{Illustration of spoofing-based evasion attacks}\label{fig:CAVVSSpoofing}
\end{figure}

Hu et al.~\cite{wei2024transferable} propose a transferable adversarial attack that generates patches trained using Intersection over Union (IoU) and classification probability loss to target object detection models, such as YOLO, in physical-world settings. The generation of adversarial perturbations is optimized by incorporating perspective transformations, spatial transformations, and similar techniques. As shown in Figure~\ref{fig:DetVehiclePatch}, the printed adversarial patch pasted on the hood of the car successfully evades detection by the YOLOv3 model.
Similarly, Oh and Yang~\cite{oh2023simulation} evaluate physical adversarial attacks in a realistic environment by creating attack textures on a 10:1 scale model of a vehicle. As shown in Figure~\ref{fig:DetVehicle}, the adversarial camouflage pattern successfully misled vehicle detectors on a customized CARLA simulator.
Sato et al.~\cite{lin2024phade} demonstrate an attack on automated lane-centering systems, widely implemented in various vehicle models equipped with Level-2 automation, which steer vehicles to maintain their position within the traffic lane. As depicted in Figure~\ref{fig:DirtyRoad}, the attack employs dirty road patterns as input perturbations to mimic natural road imperfections. The attack misleads lane detection towards the left, causing the vehicle to steer in that direction.
\begin{figure}[!ht]
    \captionsetup{format=hang,font=small, margin=5pt}
    \centering
    \subfloat[YOLOv3 detection results in indoor and outdoor scenes~\cite{wei2024transferable}\label{fig:DetVehiclePatch}]{
        \includegraphics[width=0.9\linewidth, keepaspectratio]{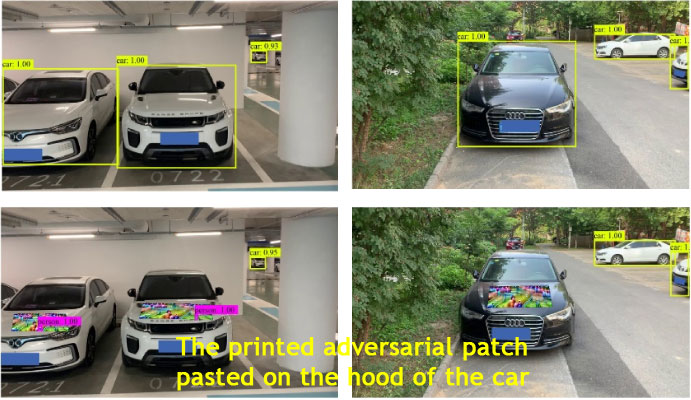}
    }\\
    \subfloat[Adversarial camouflage pattern mislead the vehicle detectors on a customized CARLA simulator~\cite{oh2023simulation}\label{fig:DetVehicle}]{
        \includegraphics[width=0.45\linewidth, keepaspectratio]{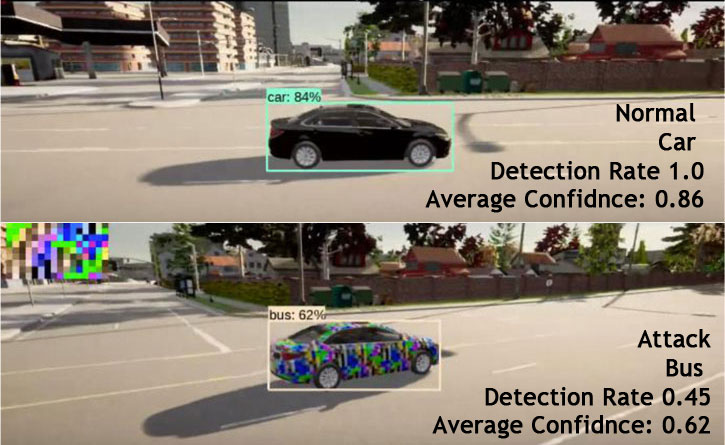}
    }
    \subfloat[Dirty road patch attack on lane detection and steering angle decisions in benign and attacked setting~\cite{lin2024phade}\label{fig:DirtyRoad}]{
        \includegraphics[width=0.45\linewidth, keepaspectratio]{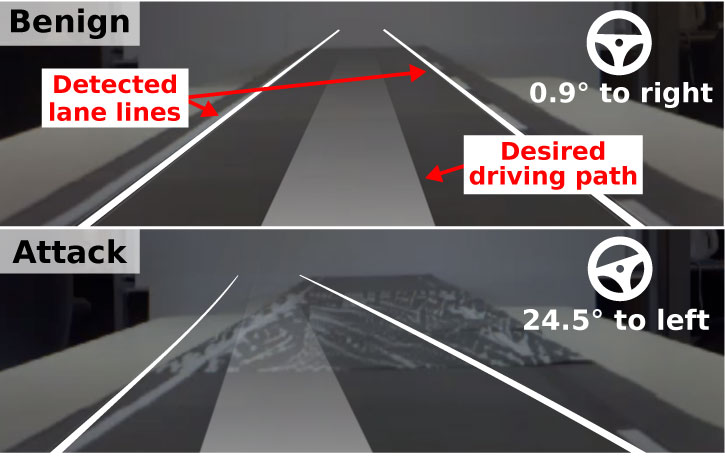}
    }
    \caption{Illustration of attacks on deep-learning-base-detector}\label{fig:CAVVSDLDetector}
\end{figure}

Traffic sign classification, including speed limits, prohibitions, instructions, and warnings, is essential for safe CAV navigation. Hsiao et al.~\cite{hsiao2024natural} investigate the effect of sunlight, headlights, and flashlight illuminations on traffic sign classification. As shown in Figure~\ref{fig:TSCNaturalLight}, the authors used a round mirror to reflect sunlight onto a 30 km/h speed limit sign, which was misclassified as 80 km/h. Similarly, Zhong et al.~\cite{zhong2022shadows} study the effect of shadows on road sign classification. Further, physical adversarial traffic signs, whether generated with imperceptible pixel changes across the entire image or by restricting visible, perceptible adversarial noise to a small area of the image, can easily deceive CAVVS classification ability~\cite{wang2023does, pavlitska2023adversarial}. 
\begin{figure}[!ht]
    \captionsetup{format=hang,font=small, margin=5pt}
    \centering
    \subfloat[A 30 km/h speed limit is misclassified as 80 km/h from natural light attack~\cite{hsiao2024natural}\label{fig:TSCNaturalLight}]{
        \includegraphics[width=0.3\linewidth, keepaspectratio]{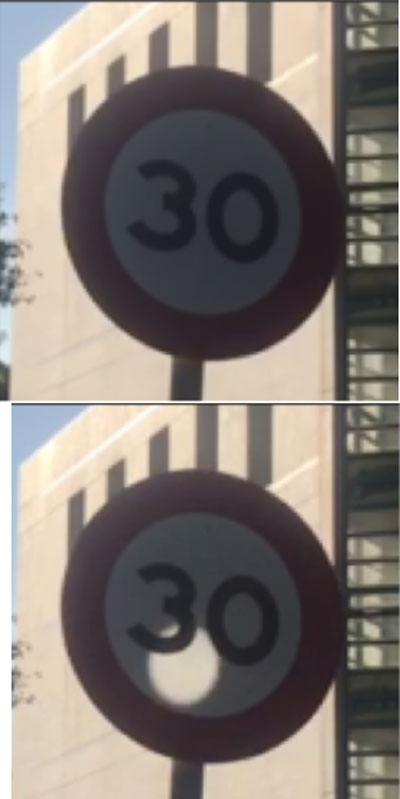}
    }
    \subfloat[GhostStripe~\cite{guo2024invisible} attack invisible to the human eye can cause traffic signs misclassification\label{fig:TSCGhostStripe}]{
        \includegraphics[width=0.3\linewidth, keepaspectratio]{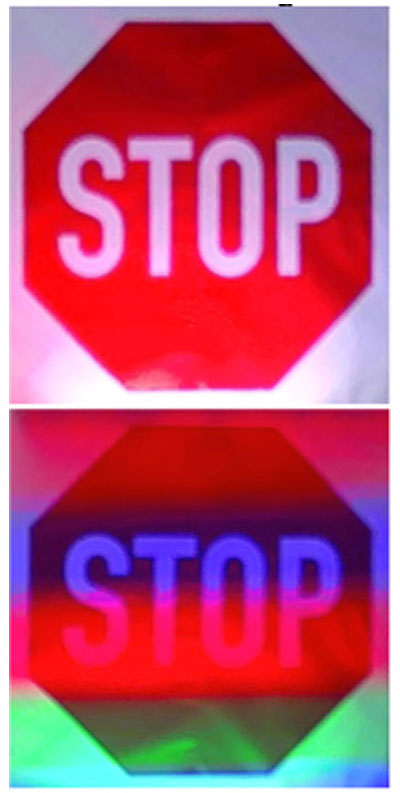}
    }
    \subfloat[Retroreflective patches~\cite{tsuruoka2024wip} are activated by the headlights of CAVs but remain dormant otherwise\label{fig:TSCRetroreflective}]{
        \includegraphics[width=0.3\linewidth, keepaspectratio]{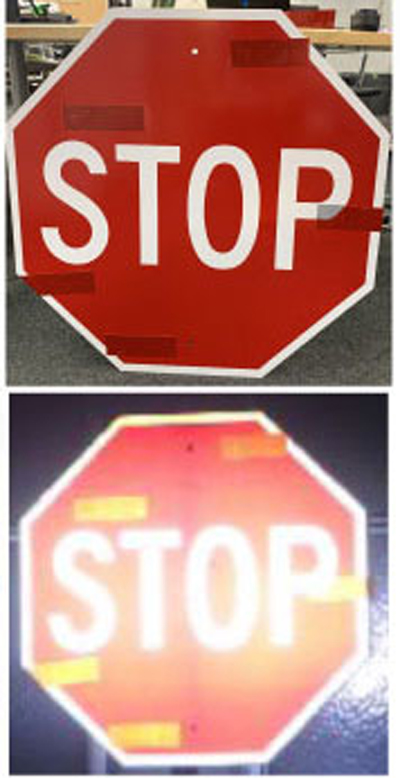}
    }
    \caption{Illustration of physical adversarial attacks in traffic sign classification}\label{fig:TrafficSignClassification}
\end{figure}

Guo et al.~\cite{guo2024invisible} employ light-emitting diodes and leverage the camera rolling shutter effect to create adversarial stripes, i.e., GhostStripe, in the captured images, as illustrated in Figure~\ref{fig:TSCGhostStripe}. According to the authors, the GhostStripe attack remains invisible to humans while consistently spoofing traffic sign recognition systems. Tsuruoka et al.~\cite{tsuruoka2024wip} design retroreflective patches to cause misidentification in traffic sign recognition tasks at nighttime. It can be observed in Figure~\ref{fig:TSCRetroreflective}, these patches are activated only by the headlights of CAVs at night, while remaining highly stealthy during the daytime, making them versatile and difficult to detect.

Studies have shown that point-cloud or 3D detectors are equally vulnerable to adversarial manipulation. Cai et al.~\cite{cai2025transferable} introduce a universal adversarial perturbation encoded as perturbation voxel units that can be added to LiDAR scenes of arbitrary scale and resolution, and that disrupts latent backbone features via a layer-activation loss to improve transferability across multiple 3D object detectors. Addressing the more restrictive hard-label setting, in which the adversary observes only the model's final output, Liu and Hu~\cite{liu2025seeing} learn a universal adversarial object trigger that can be placed naturally within a scene to mislead seven 3D models across three scene-based datasets while resisting common defenses.

Camera-based 3D and BEV (Bird's-Eye View) detectors face comparable threats. Wang et al.~\cite{wang2026invisible} propose AdvRoad, which generates road-style adversarial posters with a naturalistic appearance that blends into the road surface while causing visual 3D detectors to perceive non-existent objects at the poster's location; unlike earlier unnatural-poster attacks, AdvRoad generalizes across detectors and scenes and has been validated physically. Targeting BEV detectors specifically, Li et al.~\cite{li2026saber} propose SABER, a non-invasive attack that places 3D-consistent adversarial objects in the environment rather than on the target vehicle, using an occlusion-aware module and a BEV spatial-feature-guided optimization strategy to maintain effectiveness across camera viewpoints and time steps; the attack consistently degrades multiple BEV detectors and exposes their over-reliance on contextual scene cues. Collectively, these studies indicate that both point-cloud and camera/BEV pipelines within the CAVVS perception and data-processing layers remain vulnerable to universal, transferable, and increasingly stealthy adversarial manipulation.
}

\revisedSG{
\section{CAV vision system}\label{sec:CAVVS}
This section defines the scope of the CAV vision system (CAVVS) and presents the sensing and processing components that support environmental perception. We first describe the functional pipeline that transforms sensor observations into structured perception outputs. We then detail the key sensor inputs, their capabilities and limitations, and their role in the CAVVS.

\subsection{Scope and functional pipeline}\label{sec:CAVVSScope}
Figure~\ref{fig:CAV} illustrates a typical CAV and the placement of exteroceptive sensors. The CAVVS is the subset of the overall CAV that is responsible for perceiving the vehicle environment and providing structured information used for navigation. The CAVVS includes:

\begin{itemize}[leftmargin=*]
    \item the exteroceptive sensors that acquire observations of the     environment (cameras, LiDAR, RADAR, and ultrasonic sensors);
    \item the data-processing and machine-learning components that     transform raw sensor data into semantic representations; and 
    \item the interfaces that provide perception outputs to downstream planning and control functions.
\end{itemize}

\begin{figure}[!ht]
    \centering
    \includegraphics[width=1\linewidth]{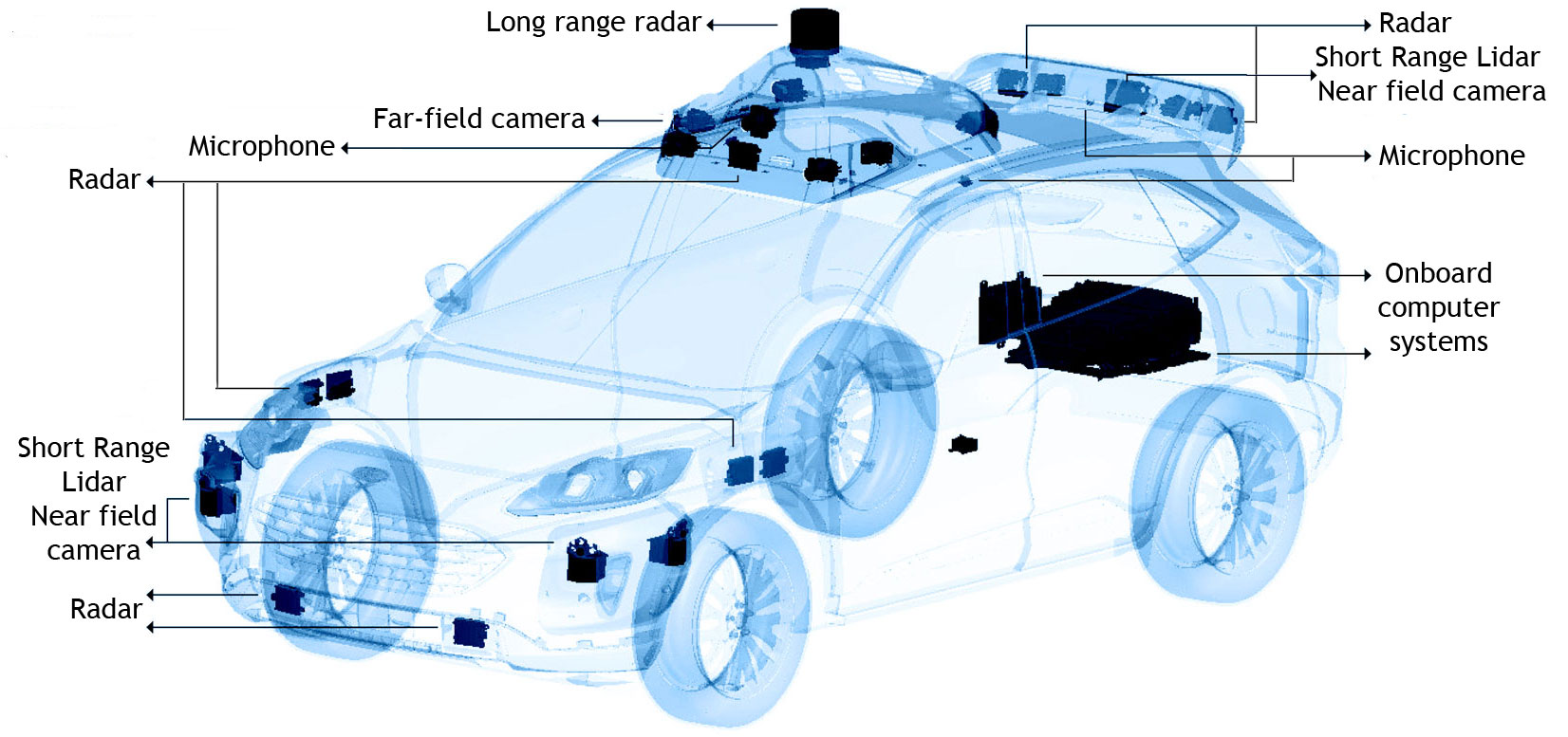}
    \caption{A typical connected and autonomous vehicle (Figure credit:~\cite{ford2023argo})}\label{fig:CAV}
\end{figure} 

The CAVVS transforms these sensor observations into structured perception outputs through a four-stage functional pipeline. Note that vehicle control, V2X communication, and other non-vision subsystems are considered only where they depend on, or influence, the CAVVS. 

\begin{enumerate}[leftmargin=*,label=(\arabic*)]
    \item \textbf{Sensing:} Cameras, LiDAR, RADAR, and ultrasonic sensors capture observations of the vehicle environment in the form of images, point clouds, range-velocity measurements, and proximity data.
    \item \textbf{Data processing:} Raw sensor data are filtered, synchronized, calibrated, and, where applicable, fused to produce consistent multi-sensor representations.
    \item \textbf{Perception and inference:} Machine-learning and rule-based algorithms perform object detection, classification, localization, tracking, and behavior prediction.
    \item \textbf{Output generation:} The results of perception are exported as structured environmental information, such as lists of detected objects with positions, velocities, and class labels, which support motion planning and control.
\end{enumerate}

This pipeline abstraction is used to organize the sensor descriptions below and to structure the reference architecture in Section~\ref{sec:SystemArch}.
Sensor fusion within the data-processing stage is commonly described in terms of early (raw-data-level), mid (feature-level), and late (decision-level) fusion~\cite{liu2023bevfusion}. BEVFusion exemplifies a mid (feature-level) design that aligns camera and LiDAR features within a unified BEV representation. This design choice affects the CAVVS attack surface: early and mid fusion combine modalities before independent verification, allowing a perturbation in one sensor stream to propagate into the shared representation, as observed in robustness benchmarks of LiDAR-camera fusion~\cite{yu2023benchmarking}. In contrast, late fusion preserves individual sensor decisions, making them easier to cross-check, but may limit the redundancy gained through joint reasoning.

\subsection{Sensor Inputs}\label{sec:SensorInputs}
Exteroceptive sensors provide the primary inputs to the CAVVS. Their role is to perceive static and dynamic objects along the CAV path and to generate data that support the stages of the functional pipeline described in Section~\ref{sec:CAVVSScope}~\cite{ghorai2022state}. The following subsections describe the key sensor modalities.

\paragraph{Camera}
Cameras provide image or video observations that are used for lane detection, object detection, traffic-sign recognition, and classification. Their outputs are high-dimensional visual data that are subsequently processed by perception models. CAVs are typically equipped with multiple cameras, including wide-angle sensors for near-field coverage and narrow-field sensors for far-field observation~\cite{he2019cooperative}. Figure~\ref{fig:CAVCamera} shows the field of view of a forward-facing camera mounted on a CAV, capturing lane markings, the lead vehicle, and the vanishing point~\cite{kim2022safety}. High-resolution image sensors detect vehicles, pedestrians, traffic lights, and road signs along the CAV path. Infrared image sensors can detect specific wavelengths in the spectral range of 0.9 to 1.7 microns, extending perception beyond the visible spectrum~\cite{sharma2022cybersecurity}.

Camera performance depends on illumination, weather, occlusion, and sensor calibration. These dependencies expose the camera to physical perturbations, image manipulation, spoofing, and adversarial inputs. For example, low-light conditions, adverse weather, and distant objects reduce classification reliability, which can be exploited by adversarial patterns or spoofed visual cues.

\begin{figure}[!ht]
    \centering
    \includegraphics[width=0.6\linewidth]{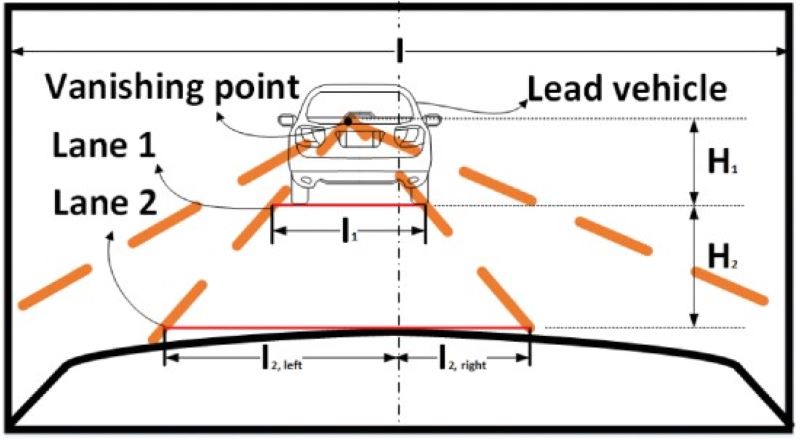}
    \caption{Field of view of a camera mounted on a CAV, capturing lane
    markings, the lead vehicle, the rear tires of the leading vehicle,
    and the vanishing point~\cite{kim2022safety}.}
    \label{fig:CAVCamera}
\end{figure}

\paragraph{LiDAR}
LiDAR (Light Detection and Ranging) sensors measure the distance to objects by timing laser pulses or measuring phase shifts, generating detailed 3-D point clouds of the vehicle surroundings~\cite{ghorai2022state, beck2023automated}. LiDAR systems are highly effective in low-light conditions and can accurately detect objects under many adverse weather conditions, although performance may degrade in heavy rain, fog, or snow. Table~\ref{tab:LiDAR} lists common LiDAR types used in CAVs. All rely on precise timing of laser pulses to compute range~\cite{foresight2023cav}.

\begin{table}[!ht]
    \centering
    \hyphenpenalty 10000
    \scriptsize
    \caption{Common LiDAR technologies for CAVs.}
    \label{tab:LiDAR}
    \begin{tabularx}{1\linewidth}{p{0.14\linewidth} p{0.19\linewidth} p{0.5\linewidth}}
        \hline
        \textbf{Type} & \textbf{Scanning method} & \textbf{Characteristics} \\\hline
        Mechanical & Rotating assembly & 360° horizontal FOV; mature technology; moving parts \\\hline
        MEMS & Micro-mirror steering & Compact; solid-state; limited FOV \\\hline
        Flash & Single-pulse illumination & No scanning; short range; robust to vibration \\\hline
        Solid-state (OPA) & Optical phased array & No moving parts; beam steering via phase control \\\hline
    \end{tabularx}
\end{table}

LiDAR outputs provide geometric structure that supports object localization, mapping, and sensor fusion. However, LiDAR systems are sensitive to cost, weather-induced attenuation, and sparse returns at long range. These characteristics expose LiDAR to point-cloud injection, laser spoofing, and manipulation of range measurements.

\paragraph{RADAR}
RADAR (Radio Detection and Ranging) sensors detect, track, and measure the distance, velocity, and angular position of objects for motion estimation~\cite{srivastav2023radars}. RADAR uses radio waves with wavelengths ranging from 30 cm to 3 mm to generate a 3-D map of the vehicle surroundings that includes both position and velocity information~\cite{beck2023automated}. RADAR can transmit and receive signals over longer distances than LiDAR and remains effective in all weather conditions, regardless of illumination or adverse weather.

Figure~\ref{fig:RadarData} illustrates FMCW radar technology used in autonomous driving, showing the detection of objects with their corresponding relative ranges, velocities, and azimuth angles~\cite{srivastav2023radars}. RADAR outputs are particularly valuable for estimating object velocity and for long-range detection. However, RADAR has lower spatial resolution than LiDAR and cameras. These limitations, together with the reliance on radio-frequency signals, expose RADAR to signal spoofing, replay, and interference.

\begin{figure}[!ht]
    \centering
    \includegraphics[width=.75\linewidth, keepaspectratio]{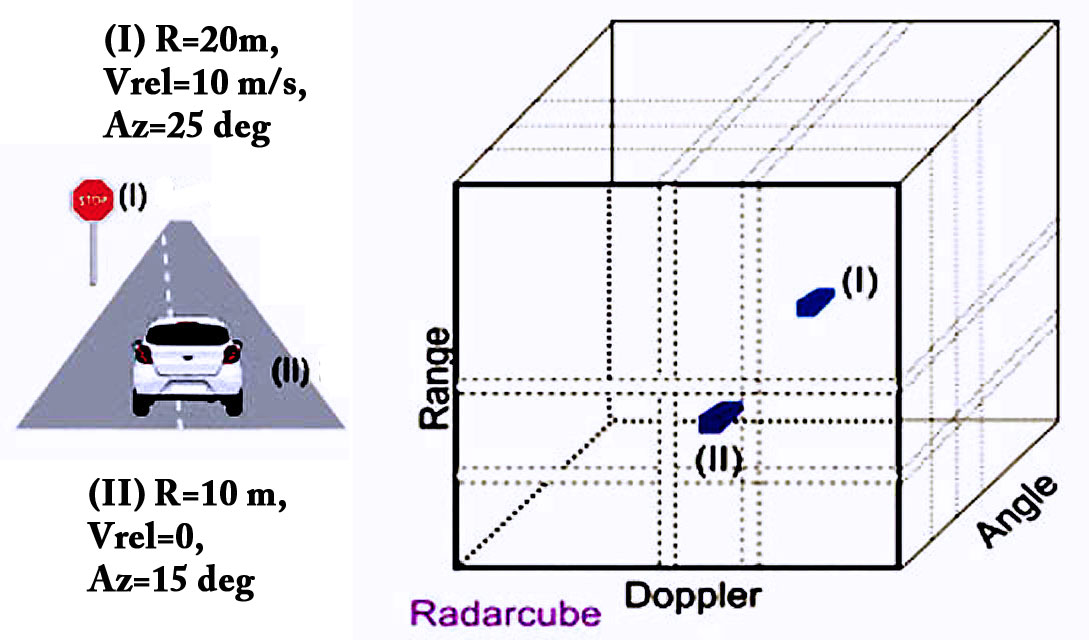}
    \caption{Illustration of FMCW radar technology used in autonomous
    driving, depicting the detection of objects with their corresponding
    relative ranges, velocities, and azimuth
    angles~\cite{srivastav2023radars}.}
    \label{fig:RadarData}
\end{figure}

\paragraph{Ultrasonic Sensors}
Ultrasonic sensors utilize sound waves to detect objects and estimate their distance from the CAV~\cite{thakur2024depth}. Also known as sonar, ultrasonic sensors complement cameras, RADAR, and LiDAR by providing a comprehensive view of nearby vehicles and obstacles~\cite{lim2018autonomous}. As shown in Figure~\ref{fig:Ultrasonic}, an ultrasonic sensor uses echolocation that is particularly effective for detecting proximity and slow-speed movements.  They also function well in fog and low-light environments, and are relatively robust to many adverse weather conditions at short range. Ultrasonic sensors are low-cost and robust for near-field sensing but provide limited semantic information and short range. Their primary security relevance lies in proximity spoofing and interference, which can affect parking assistance and low-speed maneuvering.

\begin{figure}[!ht]
    \centering
    \includegraphics[width=.7\linewidth, keepaspectratio]{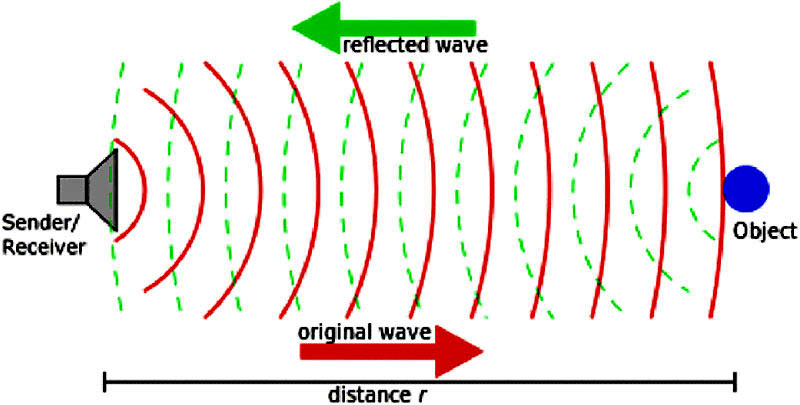}
    \caption{Ultrasonic sensor uses echolocation for detecting proximity
    and slow speeds~\cite{lim2018autonomous}.}
    \label{fig:Ultrasonic}
\end{figure}

\begin{table*}[!ht]
    \centering
    \scriptsize
\revisedSG{
    \caption{Examples of CAV vision-system sensor suites from major providers. Sensor counts and configurations are indicative of recent generations (2024-2026) and may vary by vehicle platform, market, and software configuration.}\label{tab:IndustryCAVVS}
    \begin{tabularx}{1\linewidth}{p{.12\linewidth} p{.15\linewidth} p{.15\linewidth} p{.15\linewidth} p{.32\linewidth}} \hline
    \textbf{Provider (system)} & \textbf{Cameras} & \textbf{LiDAR} & \textbf{RADAR} & \textbf{Other / notes} \\\hline
    Tesla (HW4 / ``Tesla Vision'') & 8-11 exterior cameras (wide, narrow, side, rear; 5-8 MP); cabin camera & None & HD/imaging radar & No ultrasonic sensors; vision-only perception with central FSD computer; strong emphasis on neural and temporal fusion.\\\hline
    Waymo (6th-gen Driver) & 13 high-resolution cameras (17 MP imagers; 360$\deg$ coverage) & 4 LiDAR units (roof dome + short-range; custom optics) & 6 imaging radar units (dense, temporal maps) & External audio receivers; custom 5 nm ASIC for sensor fusion.\\\hline
    Mobileye (SuperVision / Chauffeur) & 11 HD cameras (8 MP main/narrow/wing/rear; 4$\times$3 MP surround) & Front LiDAR (Chauffeur; optional depending on domain) & LR/SR radar + surround imaging radar & 2-4× EyeQ 5/6 High SoCs; camera-centric with added LiDAR/radar for higher autonomy; independent camera and radar-LiDAR subsystems for redundancy\\\hline
    Cruise (Origin / Bolt EV-based) & Multiple cameras for 360$\deg$ perception (exact count not publicly detailed) & Multiple LiDAR units (roof + peripheral) & Multiple radar units & Sensor suite between Tesla (minimal) and Waymo (dense); fused perception for urban robotaxi operations; central compute for sensor fusion and motion planning.\\\hline
    Baidu Apollo (RT6, 6th-gen) & 12 cameras (visual perception, traffic-sign recognition, 360$\deg$ coverage) & 8 LiDAR units (including blind-spot/near-range; up to 200 m range) & 6 mmWave radars & 12 ultrasonic sensors; 38-40 sensors total; up to 1,200 TOPS compute; tight integration with HD maps and Apollo perception stack\\\hline
    Zoox (Amazon robotaxi) & 14 cameras (RGB + long-wave infrared thermal sensors) & 8-12 LiDAR units (roof, bumper, side pods; 360$\deg$ coverage) & 8-20 radar modules (77 GHz; Doppler velocity up to 150 m) & Purpose-built bidirectional robotaxi; four corner sensor pods with overlapping fields of view; strong redundancy for Level 4/5-style operation; IMUs + high-precision GNSS for ego-motion. \\\hline
    Pony.ai / WeRide (representative L4 robotaxi) & 7-12 cameras (360$\deg$ coverage; traffic-sign and object perception) & 4+ LiDAR units (solid-state and/or spinning; long- and mid-range) & 4+ radar units (LR/SR) & Multi-modal suites with redundancy; configurations vary by fleet and city deployment; central compute for sensor fusion and planning. \\\hline
    Mercedes-Benz (DRIVE PILOT, L3 on S-Class/EQS) & 10+ cameras (front, rear, long-range, surround) & 1 forward-facing LiDAR (plus additional LiDAR in some descriptions) & 5+ radar units & Ultrasound sensors, rain/moisture sensors, microphones; HD maps + precision GPS; $>$30 sensors total.\\\hline
    Volvo (EX90, L3-capable with LiDAR) & 8 exterior + 2 interior cameras (in later config without LiDAR) & 1 Luminar Iris LiDAR (roof-mounted; up to 250m) & 5 radar units & 16 ultrasonic sensors; sensor fusion for Pilot Assist and safety functions. \\\hline
    \end{tabularx}
}
\end{table*}

\begin{table*}[!ht]
    \centering
    \hyphenpenalty 10000
    \scriptsize
\revisedSG{
    \caption{Comparison of key CAVVS sensor modalities.}
    \label{tab:SensorComparison}
    \begin{tabularx}{1.0\linewidth}{p{0.07\linewidth} p{0.12\linewidth} p{0.12\linewidth} p{0.12\linewidth} p{0.14\linewidth} p{0.14\linewidth} p{0.14\linewidth}}\hline
        \textbf{Sensor} & \textbf{Main output} & \textbf{Strengths} &
        \textbf{Limitations} & \textbf{Security relevance} & \textbf{Access required} & \textbf{Adversary capability} \\\hline
        Camera & Images/video & Rich semantic information & Lighting, weather, occlusion & Adversarial patches, spoofing, manipulation & Physical (line-of-sight) & Low (e.g., printed patch) \\\hline
        LiDAR & 3-D point cloud & Depth and geometry & Cost, weather sensitivity, sparse data & Point-cloud injection, laser spoofing & Physical/remote (laser) & Medium (calibrated laser) \\\hline
        RADAR & Range, velocity, angle & Long range, robust in poor visibility & Lower spatial resolution & Signal spoofing, replay, interference & Remote (RF) & Medium-high (SDR)\\\hline
        Ultrasonic & Near-field distance & Low-cost proximity sensing &
        Short range, limited semantics & Proximity spoofing, interference & Physical/near-field	& Low\\\hline
    \end{tabularx}
}
\end{table*}

To illustrate how these sensor modalities are combined in practice, Table~\ref{tab:IndustryCAVVS} summarizes commercial CAV vision-system sensor suites from major providers, showing how cameras, LiDAR, RADAR, and 
ultrasonic sensors are integrated in real-world stacks. It can be observed that providers adopt markedly different redundancy strategies for comparable driving tasks: vision‑only stacks such as Tesla's rely on 8-11 cameras with no LiDAR, whereas robotaxi platforms such as Zoox and Waymo deploy camera-LiDAR-RADAR suites. This heterogeneity across commercial platforms also complicates system integration and standardized security evaluation, since attack surfaces, redundancy assumptions, and fusion architectures vary by provider. Section~\ref{sec:OCRD} emphasizes the need for benchmark‑oriented evaluation to address cross-platform heterogeneity. Building on this, Table~\ref{tab:SensorComparison} compares the key CAVVS  sensor modalities in terms of their outputs, strengths, limitations, and security-relevant properties, including the access required and adversary capability needed to exploit each sensor. The vulnerabilities identified for 
each modality are then captured as part of the input attack surface in 
Section~\ref{sec:AttackSurfaces}.
}

\revisedSG{
\begin{figure*}[!ht]
    \centering
    \includegraphics[width=1\linewidth, keepaspectratio]{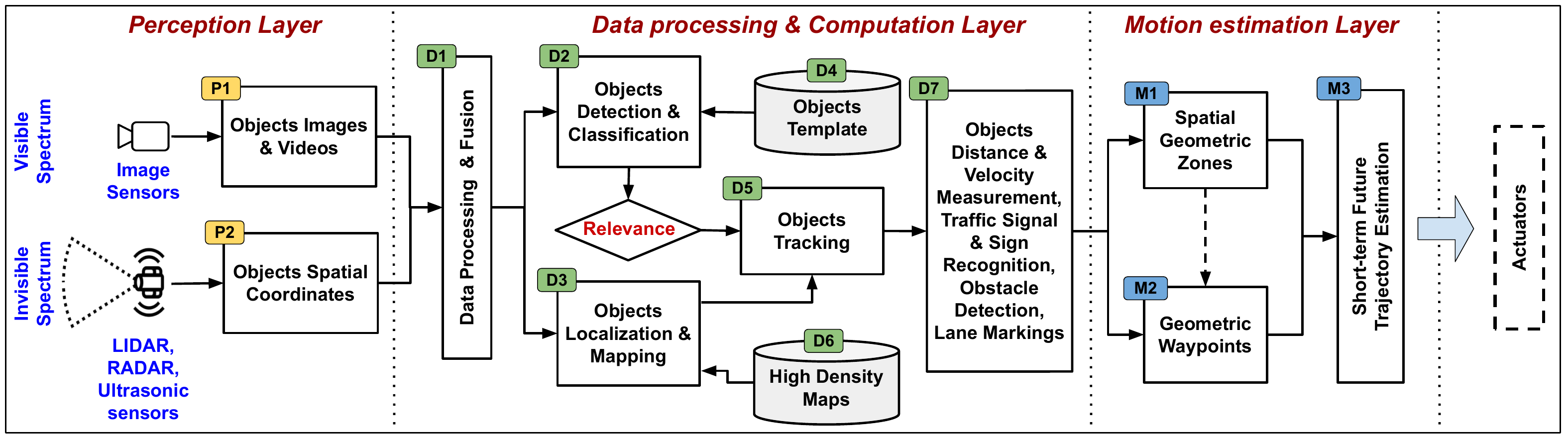}
    \caption{Reference architecture for a CAV vision system (CAVVS). The CAVVS comprises three primary layers: perception (P1, P2), data processing and computation (D1-D7), and motion estimation (M1-M3).}
    \label{fig:CAVVisionArch}
\end{figure*}

\section{CAVVS reference architecture}\label{sec:SystemArch}
Figure~\ref{fig:CAVVisionArch} presents the reference architecture proposed for the CAVVS. The architecture abstracts the system into three principal layers: the perception layer, the data-processing and computation layer, and the motion-estimation layer. The arrows indicate the primary flow of information from sensor observations to structured perception outputs and motion-related estimates. To illustrate the types of raw data that enter the perception layer, Figure~\ref{fig:SensorsData} shows example sensor outputs from a CARLA simulation, including front- and rear-facing camera images as well as LiDAR and radar measurements~\cite{beck2023automated}. These inputs correspond to the sensing stage of the functional pipeline described in Section~\ref{sec:CAVVSScope} and are processed by the components represented in Figure~\ref{fig:CAVVisionArch}.

\begin{figure}[!ht]
    \centering
    \includegraphics[width=.8\linewidth, keepaspectratio]{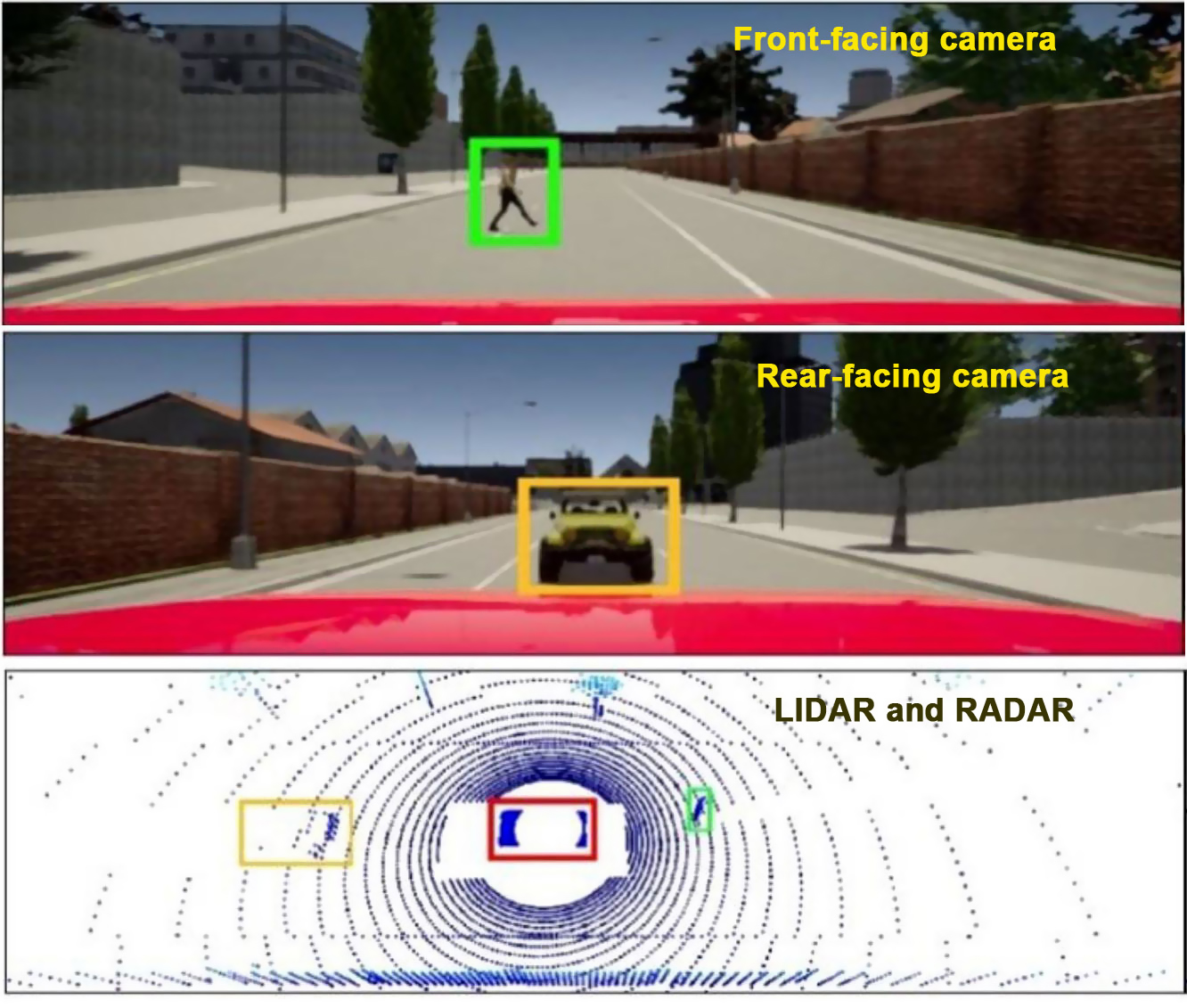}
    \caption{Illustration of sensor data in CARLA simulation~\cite{beck2023automated}. Top and middle images show front-facing and rear-facing camera data. Bottom image shows LiDAR and radar data.}\label{fig:SensorsData}
\end{figure}

\paragraph{Perception layer} 
This layer contains the exteroceptive sensors and the interfaces that
acquire observations of the vehicle environment. Its outputs include
images, point clouds, radar measurements, and proximity measurements. In
Figure~\ref{fig:CAVVisionArch}, components P1 and P2 represent the
sensing interfaces and associated front-end processing for each sensor
modality described in Section~\ref{sec:SensorInputs}. Table~\ref{tab:PerceptionLayer} presents perception-layer inputs and their use in the CAVVS pipeline.

\begin{table}[!ht]
    \centering
    \hyphenpenalty 10000
    \scriptsize
    \caption{Perception-layer inputs and their use in the CAVVS pipeline.}\label{tab:PerceptionLayer}
    \begin{tblr}{
        width=1\linewidth,
        colspec = {p{.1\linewidth} p{.28\linewidth} p{.48\linewidth}},
    }\hline
    \textbf{Sensor} & \textbf{Data representation} & \textbf{Primary perception uses}\\\hline
    Camera & 2D image/video frames (RGB, IR) &
    Lane detection, traffic-sign and light recognition, semantic segmentation,
    2D/3D object detection and classification~\cite{ghorai2022state}\\\hline
    LiDAR & 3D point clouds (x,y,z, intensity, possibly reflectivity) &     3D object detection and localization, mapping, depth estimation,     scene geometry reconstruction~\cite{beck2023automated}\\\hline
    RADAR & Range-angle-Doppler measurements (range, azimuth, velocity) &    Long-range object detection, velocity estimation, tracking in adverse    weather~\cite{thakur2024depth}\\\hline
    Ultrasonic & Scalar distance measurements (time-of-flight) & Near-field obstacle detection, parking assistance, low-speed    maneuvering~\cite{lim2018autonomous}\\\hline
    \end{tblr}
\end{table}
    
\paragraph{Data processing and computation layer} 
This layer performs calibration, synchronization, filtering, sensor fusion, feature extraction, machine-learning inference, and storage or exchange of relevant data. Components D1-D7 in Figure~\ref{fig:CAVVisionArch} correspond to processing stages such as preprocessing, object detection, classification, localization, tracking, behavior prediction, and data management. This layer transforms raw sensor data into semantic representations of the environment.

Data fusion~\cite{thakur2024depth} can be performed at three levels: low-level (raw data), mid-level (feature extraction), and high-level (decision-making), corresponding respectively to the early, mid, and late fusion categories discussed in Section~\ref{sec:CAVVSScope}. Mid-level fusion is commonly realized through a unified BEV representation (Section~\ref{sec:CAVVSScope}), which several components in this layer consume as a shared input for object detection, classification, and tracking. The object classification module determines the relevance of objects for tracking and the identification of the state of traffic signals. Machine learning (ML) is applied to transform, detect, and segment visual data to extract information for CAV navigation. Subsequently, the object tracking module initiates the measurement of distance and velocity of objects of interest using 3D point clouds provided by the object localization and mapping module~\cite{beck2023automated}. 
    
\paragraph{Motion estimation layer} 
This layer transforms perception outputs into estimates of object
position, velocity, trajectory, and scene dynamics. Components M1-M3 in
Figure~\ref{fig:CAVVisionArch} generate motion-related estimates that
support planning and control functions. The sensor data processed by the previous layer is exploited to determine a safe trajectory for CAV navigation~\cite{teng2023motion}. The safe trajectory can be defined as a sequence of spatiotemporal waypoints in free space, represented by state variables such as position, orientation, and their corresponding rates of change: linear velocity, linear acceleration, angular velocity, and angular acceleration. 

The short-term future trajectory (STFT) is calculated based on the free space, i.e., \textit{spatial geometric zones}, and the corresponding sequence of space-related states, i.e., \textit{geometric waypoints}. Based on the STFT, the actuators decide whether to steer, accelerate, or brake the CAV.  Table~\ref{tab:MotionEstimationLayer} summarizes how the motion-estimation layer maps the planned STFT to actuator commands for throttle, steering, and braking, which realize the CAV's longitudinal and lateral motion.

\begin{table}[!ht]
    \centering
    \hyphenpenalty 10000
    \scriptsize
    \caption{Mapping from motion-estimation outputs to actuator commands.}
    \label{tab:MotionEstimationLayer}
    \begin{tblr}{
        width=1\linewidth,
        colspec = {p{.1\linewidth} p{.8\linewidth}},
    }\hline
    \textbf{Actuator} & \textbf{Role in executing the planned trajectory}\\\hline
    Throttle & Converts desired longitudinal acceleration from the motion-estimation layer into throttle commands, regulating engine or motor torque to achieve target speed and spacing along the planned trajectory.\\\hline
    Steering & Translates desired path curvature and lane-change maneuvers into steering-angle commands, controlling lateral motion to follow the
    geometric waypoints of the short-term future trajectory.\\\hline
    Brakes & Applies deceleration commands to reduce speed or bring the CAV to a stop, ensuring compliance with safety constraints and obstacle
    avoidance requirements derived from the estimated trajectory.\\\hline
    \end{tblr}
\end{table}

}

\revisedSG{
\section{Threat model and attack-surface derivation}\label{sec:TMASD}
This section defines the threat model for the CAVVS by identifying its attack surfaces based on the reference architecture and system lifecycle, addressing \textbf{RQ2} and \textbf{RQ3}. We first establish the scope and assumptions of the threat model, specifying the adversary's objectives and capabilities as well as the assets to be protected. We then use a baseline computer-vision system to map known attack points to the CAVVS modules defined in Section~\ref{sec:SystemArch}. Finally, we consolidate these findings into three CAVVS attack surfaces (data, models, and inputs; secondary). Outputs and interfaces are treated as secondary asset classes whose vulnerabilities are realized through these surfaces. These surfaces provide the basis for the attack-vector analysis presented in Section~\ref{sec:AV}.

\subsection{Threat-model scope and assumptions}\label{sec:ThreatModel}
We consider attacks across the design, development, testing, deployment, and inference phases of the CAVVS lifecycle. The protected assets include sensor inputs, training and testing data, learned models, model parameters, processing logic, perception outputs, and interfaces between CAVVS components. Table~\ref{tab:CAVVSAssets} summarizes the main asset classes considered in the threat model. 

\begin{table}[!ht]
    \centering
    \scriptsize
    \caption{CAVVS assets considered in the threat model.}\label{tab:CAVVSAssets}
    \begin{tabularx}{1\linewidth}{p{.1\linewidth} X}\hline
        \textbf{Asset} & \textbf{Examples} \\\hline
        Inputs & Camera images, LiDAR point clouds, RADAR measurements, ultrasonic measurements, and received messages. \\
        Data & Training, validation, testing, calibration, and operational data. \\
        Models & Model architecture, parameters, weights, configurations, and inference logic. \\
        Outputs & Object labels, locations, velocities, trajectories, and confidence scores. \\
        Interfaces & Sensor, storage, communication, software, and actuator interfaces. \\
        \hline
    \end{tabularx}
\end{table}

The adversary is assumed to have physical, remote, or privileged access, depending on the attack under consideration. Physical access enables the manipulation of objects in the vehicle environment or interference with exposed sensors. Remote access enables the manipulation of communicated data or signals. Privileged access enables modification of stored data, model parameters, software, or system configurations. We do not assume that the adversary possesses all these capabilities simultaneously. Instead, each attack is evaluated based on the specific access and capabilities required for its execution. Operational conditions such as weather (e.g., rain, snow, fog), illumination, and occlusion may degrade sensor performance and influence attack effectiveness~\cite{zhang2023perception}. These conditions are treated as environmental factors that can amplify or mask the effects of an attack rather than as attacks themselves.

Real-world demonstrations, such as manipulated traffic signs~\cite{yan2023adversarial, cao2021invisible, poddar2024comprehensive}, fabricated obstacles via LiDAR spoofing and mirror-based attacks~\cite{cao2021invisible, tu2020physically}, and adversarial patterns on pedestrians' clothing~\cite{yan2023adversarial, nassi2020phantom, serban2020adversarial}, have shown that such attacks can exploit vulnerabilities in the CAVVS assets listed in Table~\ref{tab:CAVVSAssets}, which arise from their functional roles and interfaces~\cite{yan2023adversarial}. In particular, sensor inputs are susceptible to spoofing, tampering, and adversarial perturbations. Spoofing attacks can present falsified or manipulated images and videos or interfere with radar and LiDAR signals~\cite{poddar2024comprehensive}. Tampering attacks can modify traffic signs using stickers or graffiti or disrupt LiDAR sensors through laser interference~\cite{cao2021invisible}. Adversarial perturbations can be introduced through geometric transformations or generative noise~\cite{serban2020adversarial, yan2023adversarial}.

Data assets are susceptible to unauthorized access, manipulation, and exfiltration. Model assets are susceptible to extraction, parameter manipulation, and logic corruption. Output assets are susceptible to misclassification, falsification, and denial of service. Component interfaces are susceptible to MitM, replay, and injection attacks. The exploitability of these vulnerabilities depends on the adversary's access, the lifecycle phase, and the targeted CAVVS module. However, outputs and interfaces are not treated as distinct attack surfaces. Their vulnerabilities are instead captured by the three attack surfaces derived below and the attack vectors analyzed in Section~\ref{sec:AV}. Specifically, interface compromises are captured as input-targeting MitM (AV8), while output falsification results from the evasion (AV7) and logic-corruption (AV4) vectors targeting inputs and models, respectively.

\subsection{Attack-surface derivation}\label{sec:ASD}
Because a CAVVS is a sensor-based machine-learning system, its attack surfaces can be analyzed using concepts established for other sensor-based recognition systems. We use a generic biometric recognition system (BRS) as a methodological baseline because it provides a well-established decomposition of sensing, feature extraction, model processing, matching, and decision functions~\cite{joshi2020comprehensive, gupta2023survey}. The baseline is not intended to imply that biometric and automotive systems are identical. Instead, it provides a systematic set of attack points that can be mapped to the functionally corresponding CAVVS modules.

\paragraph{Baseline computer-vision system}\label{sec:BRS}
A computer vision system typically involves image acquisition through sensors, followed by data processing and analysis using machine learning algorithms~\cite{ellis2024machine}. Figure~\ref{fig:BRS} illustrates the mapping of various modules in the perception and data-processing layers of the CAVVS architecture depicted in Figure~\ref{fig:CAVVisionArch}. We use this mapping to analyze how the sixteen extensively studied attack points (APs) in the BRS relate to the modules in the CAVVS for attack-surface determination.

\begin{figure}[!hpt]
    \centering
    \includegraphics[width=1\linewidth, keepaspectratio]{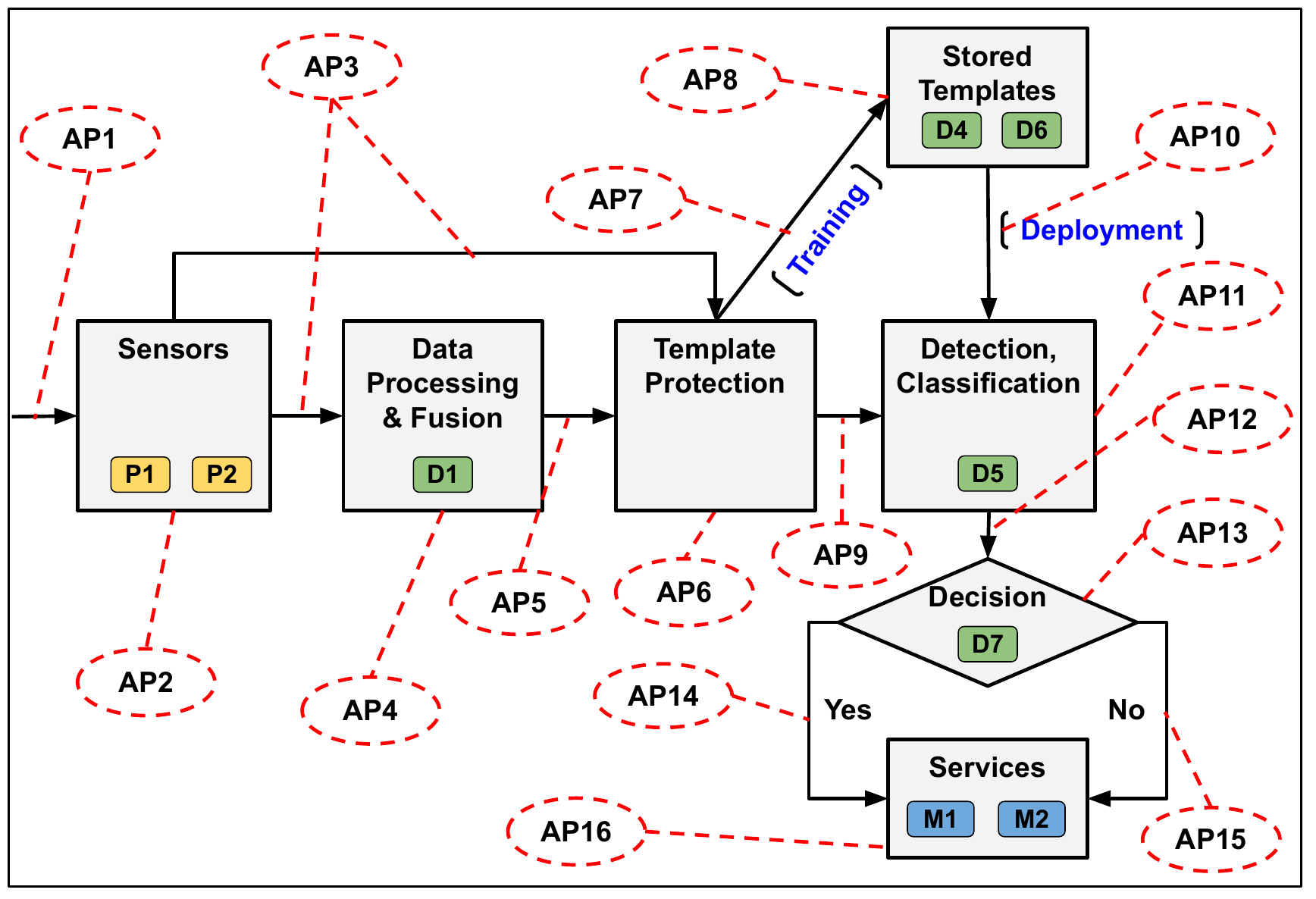}
    \caption{Attack points (AP) in a generic BRS~\cite{gupta2023survey}. [AP1 - Adversary attacks (spoofing, presentation, fake physical biometrics, latent print reactivation, wolf attack), Administrative frauds (false enrollment, exception abuse); AP2 - Denial of service; AP3 - Man-in-the-middle attack, replay attack, Hill-climbing, brute force attack, dictionary attack; AP4 - Override feature extractor, Trojan horse attack; AP5 - False data inject, reuse of residual; AP6 - Side channel attack; AP7 - Intercept template, data reject; AP8 - Unauthorized access through external compromised system, Steel, delete, modify, substitute, reconstruct templates; AP9 - Alter system parameters, Synthesized feature vector, brute force attack, hill-climbing, fake digital biometrics, replay attack; AP10 - Intercept stored template, replay attack; AP11 - Comparison module override, side channel attack, Trojan horse attack; AP12 - Modify score;  AP13 - Override decision module; AP14 - False match; AP15 - False non-match; AP16 - Denial of service]}\label{fig:BRS}
\end{figure}

Subsequently, by analyzing the attack points shown in Figure~\ref{fig:BRS}, we note that each attack targets the system data, inputs, or models with the intent of compromising CIA properties. Table~\ref{tab:ThreatModel} categorizes each attack point, highlighting the phases (i.e., training, testing, and inference) and potential attack surfaces.
\begin{table}[!ht]
    \centering
    \scriptsize
    \caption{Illustration of attack points, phase-wise impact, and attack surfaces\label{tab:ThreatModel}}
    \begin{tabularx}{1\linewidth}{|c|p{.5\linewidth}|C|C|}\hline
    & \textbf{Attack points} & \textbf{Phase} & \textbf{Asset}\\\hline
    \multirow{2}{*}{AP1} & Spoofing, presentation, fake physical biometrics, latent print reactivation, wolf attack & Inference & Input\\\cline{2-4}
    & False enrollment, exception abuse & Training, Testing & Data\\\hline
    AP2	& Denial of service & Inference & Input\\\hline
    AP3	& Hill-climbing, brute force attack, dictionary attack,
    replay attack, Mitm attack & Inference & Input\\\hline
    AP4	& Override feature extractor, Trojan horse attack &	Inference & Model\\\hline
    AP5	& False data inject, reuse of residue & Inference & Input\\\hline
    AP6	& Side channel attack & Inference & Model\\\hline
    AP7	& Intercept template, data reject &	Training, Testing & Data\\\hline
    AP8	& Unauthorized access through external compromised system, Steel, delete, modify, substitute, reconstruct templates &	Training, Testing & Data, Model\\\hline
    AP9	& Alter system parameters, Synthesized feature vector, brute force attack, hill-climbing, fake digital biometrics, replay attack & Inference &	Input, Model\\\hline
    AP10 & Intercept stored template, replay attack	& Inference	& Data\\\hline
    AP11 & Comparison module override, side channel attack, Trojan horse attack & Inference	& Model\\\hline
    AP12 & Modify score	& Inference & Model\\\hline
    AP13 & Override decision module & Inference & Model\\\hline
    AP14 & False match & Inference & Model\\\hline
    AP15 & False non-match & Inference & Model\\\hline
    AP16 & Denial of service & Inference & Model\\\hline
    \end{tabularx}
\end{table}

\paragraph{Mapping attack points to the CAVVS}
The mapping in Table~\ref{tab:ThreatModel} shows that the biometric attack points correspond primarily to three CAVVS asset classes: inputs, data, and models. \textit{Input attacks} affect observations presented to the perception pipeline, \textit{data attacks} affect training, testing, calibration, or stored operational data, and \textit{model attacks} affect learned parameters, processing logic, or inference decisions. This mapping motivates the three CAVVS attack surfaces defined in the following subsection.

\subsection{CAVVS Attack Surfaces}\label{sec:AttackSurfaces}
The mapping in Table~\ref{tab:ThreatModel} yields three major CAVVS attack surfaces: data, models, and inputs. In response to \textbf{RQ3},  Figure~\ref{fig:CAVVSLifecycle} summarizes their relationship with the CAVVS lifecycle. The \emph{data attack surface} includes training, validation, testing, calibration, and operational data, as well as the repositories and interfaces through which these data are collected, processed, and stored. The \emph{model attack surface} includes model architectures, parameters, weights, inference logic, configurations, and execution environments. The \emph{input attack surface} includes sensor observations and data received through external interfaces during inference.

\begin{figure}[!ht]
    \centering
    \includegraphics[width=1.0\linewidth, keepaspectratio]{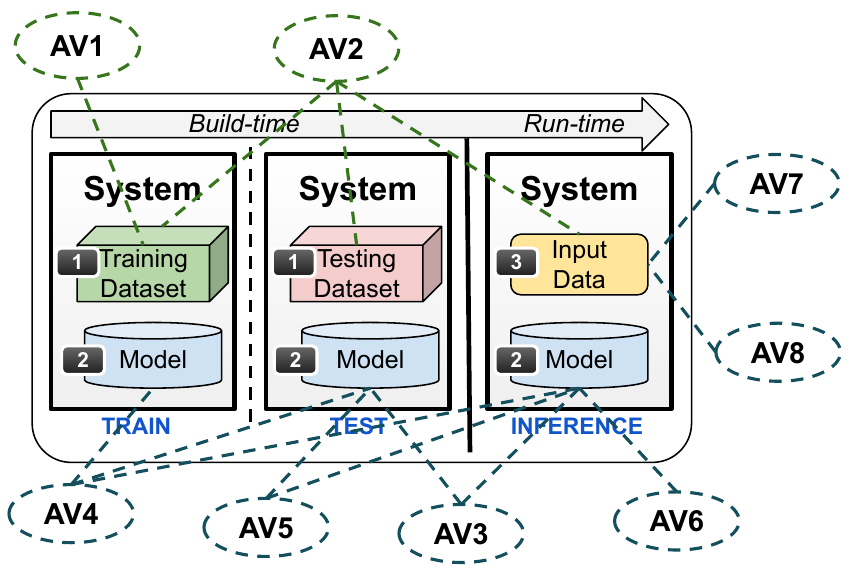}
    \caption{Illustration of vision system life cycle and attack surfaces 1) data, 2) models, and 3) inputs. [AV1: Data poisoning; AV2: Data exfiltration; AV3: Model extraction; AV4: Logic corruption; AV5: Membership inference; AV6: Side channel; AV7: Evasion; AV8: Man-in-the-middle]}\label{fig:CAVVSLifecycle}
\end{figure}

These surfaces are analytically distinct but interconnected. For example, a data-poisoning attack can compromise the data surface during training and subsequently alter the model surface. Similarly, an input manipulation attack can exploit the input surface during inference and cause an incorrect model output. This relationship motivates the organization of Section~\ref{sec:AV}, where attack vectors are analyzed according to the surface they primarily target. Data poisoning~\cite{wang2022poisoning} and data exfiltration~\cite{sabir2021machine} primarily target the data surface. Model extraction~\cite{oliynyk2023know}, logic corruption~\cite{jeong2020artificial}, membership inference~\cite{rezaei2021difficulty}, and side-channel~\cite{hua2022reverse} attacks primarily target the model surface. Evasion~\cite{wang2023does} and MitM~\cite{wang2020man} attacks primarily target the input surface, although some attacks could affect more than one surface.
}

\revisedSG{
\section{Attack vectors}\label{sec:AV}
Addressing \textbf{RQ4} and \textbf{RQ5}, this section examines the attack vectors for each identified attack surface illustrated in Figure~\ref{fig:CAVVSLifecycle}, which is a simplified abstraction of the CAVVS reference architecture (refer to Figure~\ref{fig:CAVVisionArch}). We further discuss the strategies and mechanisms that the adversary can employ to generate attack vectors. Understanding the characteristics of these attack vectors is crucial for implementing effective security solutions to enhance the robustness of CAVVS.

\subsection{Attacks targeting data}\label{sec:AttacksData}
As illustrated in Figure~\ref{fig:CAVVisionArch}, a robust CAVVS requires a secure data pipeline (\YC{P1}, \YC{P2}, \GC{D4}, \GC{D6}) spanning data collection, knowledge curation, and storage~\cite{bhardwaj2024machine}. This pipeline transforms the sensor inputs described in Section~\ref{sec:SensorInputs} into the training, testing, calibration, and operational data that collectively constitute the data attack surface introduced in Section~\ref{sec:AttackSurfaces}. As accurate perception is essential for safe and secure driving, the quality and integrity of these data play a critical role in determining perception performance~\cite{wang2024data}. Table~\ref{tab:AttacksData} lists attacks targeting data, their impact on CIA properties, and the strategies or mechanisms used to accomplish them.
\begin{table}[!ht]
    \centering
    \scriptsize
    \hyphenpenalty=10000
    \caption{Attacks targeting the data surface.
    \textit{D} = direct impact; \textit{ID} = indirect impact; -- = no identified impact.}
    \label{tab:AttacksData}
    \begin{tblr}{
        width=\linewidth,
        colspec={ |p{.04\linewidth}|p{.15\linewidth}|p{.04\linewidth}|p{.04\linewidth}|p{.04\linewidth}|p{.4\linewidth}|
        },
        row{1}={c},
        row{2}={c},
    }
    \hline
    & & \SetCell[c=3]{c}{\textit{Security property}} & & & \\\cline{3-5}
    \# & \textbf{Attack type} & \textbf{C} & \textbf{I} & \textbf{A}
    & \textbf{Mechanism/strategy} \\\hline
    AV1 & Data poisoning & -- & \textit{D} & -- &
    Injection, deletion, alteration, label flipping, adversarial examples,
    backdoor insertion, MitM, and manipulation of data pipelines \\\hline
    AV2 & Data exfiltration & \textit{D} & -- & -- &
    Phishing, malware, privilege escalation, compromised storage or APIs,
    insider access, and side-channel leakage \\\hline
    \end{tblr}
\end{table}

\subsubsection{Data poisoning (AV1)}
\noindent\textbf{Mechanism/strategy:} Data poisoning compromises the integrity of the vision system by manipulating training data through injection, deletion, or alteration~\cite{zhou2022adversarial}, and can be orchestrated via adversarial examples introduced into the training set~\cite{clifford2022autonomous} or via MitM control of the data pipeline during collection~\cite{wang2024data}. Dirty-label variants, including label flipping, manipulate classifiers to misclassify specific inputs, such as a stop sign carrying an attached sticker, leading to decreased accuracy and biased or underperforming models. Similarly, backdoor insertion, embeds a malicious trigger into the training data so that the poisoned model behaves normally on benign inputs but misclassifies on demand when the trigger is present. 

Formally, a poisoning adversary seeks a perturbed training set $D' = D \cup \Delta$, where $\Delta$ denotes injected, altered, or mislabeled samples bounded by a budget $|\Delta| \leq \epsilon$, so as to maximize the trained model's loss on a target distribution (availability attacks) or on a specific subset of inputs (targeted or subpopulation attacks), while keeping $D'$ statistically close enough to $D$ to evade data-validation checks~\cite{fan2022survey}. This budget-constrained framing clarifies why availability attacks typically require broader data access than targeted attacks, which can succeed with a small number of carefully chosen samples. Poisoned datasets can distort traffic data, causing suboptimal routing decisions and reduced network efficiency~\cite{wang2024data, wang2022poisoning}. Similarly, poisoning LiDAR data can lead to the misclassification or misinterpretation of road objects.

\noindent\textbf{Targeted layer(s):} Perception layer (\YC{P1}, \YC{P2}), where raw sensor data is acquired, and the data-processing and computation layer (\GC{D4}, \GC{D6}), where data is fused, curated, and stored (refer Section~\ref{sec:SystemArch}).

\subsubsection{Data exfiltration (AV2)}
\noindent\textbf{Mechanism/strategy:} Data exfiltration is the unauthorized removal or theft of data from a system, often resulting in privacy violations, loss of intellectual property, regulatory breaches, and substantial financial harm~\cite{sabir2021machine}. Adversaries can extract sensitive data from CAVVS, including training data, trained weights, and proprietary algorithms, impacting system confidentiality. Common mechanisms include phishing, malware, privilege escalation, and side-channel attacks, often resulting from insider attacks, hacktivism, unauthorized access, human error, or compromised storage/APIs~\cite{chung2020interactive, gupta2019risk}. Authorized individuals, including employees and contractors, can intentionally extract data for personal gain, while CAVs can also be infiltrated through unauthorized access, hacking techniques, or malicious software to extract data undetected.

\noindent\textbf{Targeted layer(s):} Primarily the data-processing and computation layer (\GC{D6}), where data is stored and curated, together with the storage and communication interfaces through which data is accessed (refer Section~\ref{sec:SystemArch}).

\subsection{Attacks targeting a model}
Referring to the CAVVS reference architecture in Figure~\ref{fig:CAVVisionArch}, model-directed attacks target the learned models and processing logic deployed in the data-processing and computation layer, particularly modules \GC{D1}, \GC{D5}, and \GC{D6}. These modules contains different neural-network architectures depending on the perception task, including object detection, temporal prediction, feature extraction, anomaly detection, sensor fusion, and contextualization~\cite{chib2023recent, gupta2025investigation}.

Figure~\ref{fig:VisionDDNs} provides an overview of the deep-neural-network (DNN) families that support above functionalities. Convolutional Neural Networks (CNNs), You Only Look Once (YOLO), Single Shot MultiBox Detector (SSD), and Faster Region-based Convolutional Neural Networks (Faster R-CNNs) can support image-based object detection~\cite{teng2023motion, beck2023automated, ghorai2022state, girdhar2023cybersecurity}. Recurrent Neural Networks (RNNs) and Long Short-Term Memory (LSTM) networks can support temporal modelling and obstacle-behaviour analysis~\cite{sharma2019attacks}. Autoencoders (AEs) and Deep Belief Networks (DBNs) can support feature transformation, reconstruction, denoising, and anomaly detection~\cite{abdo2023cybersecurity}. Siamese Neural Networks (SNNs), Transformers, and Vision Transformers (ViTs) can support object matching, tracking, contextualization, and trajectory prediction~\cite{thakur2024depth}. Generative Adversarial Networks (GANs) and diffusion models can support data augmentation and image or scene generation~\cite{thakur2024depth, sai2024dmdat}. 
\begin{figure}[!ht]
    \centering
    \includegraphics[width=1\linewidth, keepaspectratio]{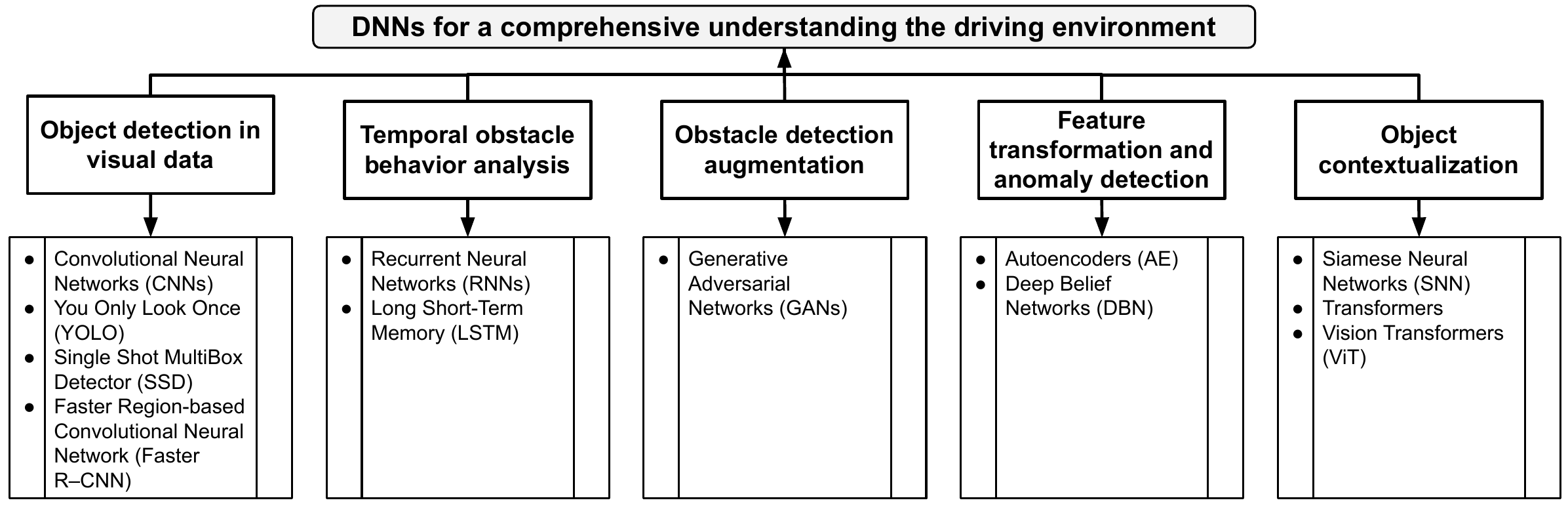}
    \caption{Illustration of DNNs used in CAVs vision system for a comprehensive understanding of the driving environment~\cite{thakur2024depth}}\label{fig:VisionDDNs}
\end{figure}

The figure illustrates the diversity of model components that can be incorporated into the CAVVS processing pipeline. This diversity expands the model attack surface: the adversary can target the architecture, parameters, learned representations, inference logic, training or fine-tuning process, or the information exposed through the model interface. Table~\ref{tab:AttacksModel} lists attack vectors targeting the model surface and relates each vector within the CAVVS pipeline, the adversary capability required, and the resulting CIA impact. 

\begin{table}[!ht]
    \centering
    \scriptsize
    \hyphenpenalty=10000
    \caption{Attacks targeting the model surface. \textit{D} = direct impact; \textit{ID} = indirect impact; -- = no identified impact.}\label{tab:AttacksModel}
    \begin{tblr}{
        width=\linewidth,
        colspec={
            |p{.04\linewidth}|p{.15\linewidth}|p{.04\linewidth}| p{.04\linewidth}|p{.04\linewidth}|p{.4\linewidth}|
        },
        row{1}={c},
        row{2}={c},
    }\hline
    & & \SetCell[c=3]{c}{\textit{Security property}} & & & \\\cline{3-5}
    \# & \textbf{Attack type} & \textbf{C} & \textbf{I} & \textbf{A}
    & \textbf{Mechanism/strategy} \\\hline
    AV3 & Model extraction & \textit{D} & -- & -- &
    Query-based replication, knowledge distillation, reverse engineering,
    unauthorized access to model artefacts, and side-channel observation \\
    \hline
    AV4 & Logic corruption & -- & \textit{D} & -- & Trojan model updates, parameter tampering, model drift or decay, and compromised training or fine-tuning pipelines \\\hline
    AV5 & Membership inference & \textit{D} & -- & -- & Repeated querying, confidence analysis, output inspection, and exploitation of overfitting \\\hline
    AV6 & Side-channel attacks & \textit{D} & \textit{ID} & -- & Timing analysis, cache observation, power analysis, electromagnetic leakage, and execution-trace analysis \\\hline
    \end{tblr}
\end{table}

\subsubsection{Model extraction (AV3)}
\noindent\textbf{Mechanism/strategy:} Model extraction~\cite{oliynyk2023know} refers to the theft of a model architecture, training hyperparameters, learned parameters, or its behavioral approximation, thus compromising model confidentiality, as summarized in Table~\ref{tab:AttacksModel}. The adversary aims to replicate a target model's functionality by constructing a similar model with comparable predictive performance~\cite{hu2022membership}; these attacks are also referred to as inference attacks, exploratory attacks, copying, duplication, mimicking, or model approximation. Knowledge distillation, querying, reverse-engineering, and unauthorized access are common strategies: knowledge distillation trains a smaller model by transferring knowledge from a larger, pre-trained network; reverse engineering analyzes the hardware or software characteristics of a CAVVS; and unauthorized access exploits cache or timing side channels.

\noindent\textbf{Targeted layer(s):} Data-processing and computation layer (\GC{D1}, \GC{D5}, \GC{D6}), where model architecture, parameters, and inference logic reside (refer Section~\ref{sec:SystemArch}).

\subsubsection{Logic corruption (AV4)}
\noindent\textbf{Mechanism/strategy:} As summarize in Table~\ref{tab:AttacksModel}, logic corruption is the deliberate alteration of model parameters or structural components to induce incorrect or malicious behavior in a CAVVS~\cite{wang2022poisoning}. For instance, adversarial traffic signs present in training data can compromise a CAVVS's ability to accurately classify signs during real-time driving~\cite{jiang2020poisoning}. As discussed under AV1, data-poisoning attacks that lead to logic corruption fall into availability, targeted, and subpopulation categories~\cite{fan2022survey}. Availability attacks can maximize average test loss to degrade overall performance. Targeted attacks can cause misclassification on one or a small number of test samples. And, subpopulation attacks can degrade performance on specific subgroups while preserving overall accuracy. Backdoor attacks~\cite{han2023interpreting} embed malicious triggers in the training data so that the model behaves normally on benign inputs yet misclassifies when the trigger is present at inference time constituting a specialized instance of data poisoning~\cite{jiang2020poisoning}.

Related non-adversarial threats to logic integrity include concept drift~\cite{suarez2023survey} and model decay~\cite{nayak2021concept}. Concept drift arises from unexpected changes in the data distribution over time, to which CAVVS is particularly exposed given its reliance on multiple sensors; model decay reflects performance degradation irrespective of data-distribution changes. Although these phenomena are not adversarial attacks, they can produce effects similar to logic corruption and therefore require consideration when evaluating the long term robustness of CAVVS models.

\noindent\textbf{Targeted layer(s):} Data-processing and computation layer (\GC{D1}, \GC{D5}), where classification and inference logic execute (refer Section~\ref{sec:SystemArch}).

\subsubsection{Membership inference (AV5)}
\noindent\textbf{Mechanism/strategy:} As shown in Table~\ref{tab:AttacksModel}, membership inference attacks aim to determine whether a specific data point was included in a model's training set, compromising confidentiality by potentially exposing sensitive training information~\cite{carlini2022membership}. The adversary builds an attack model that uses the target model's output (confidence values) to infer membership status~\cite{shokri2017membership}, typically using shadow models trained on known training/non-training samples. Success relies on structural similarity between shadow and target models and comparability of their training data distributions. The adversary can exploit intermediate layer activations, input and weight gradients, and decision-boundary distances to mount such attacks on naturally trained models~\cite{rezaei2021difficulty}.
Adversarially trained models are particularly vulnerable to membership inference, with privacy leakage increasing with model robustness~\cite{song2019membership}. This holds across both empirical and provable/certified adversarial defenses, indicating that the privacy cost is not specific to any one defense strategy but a broader trade-off between robustness and privacy.

\noindent\textbf{Targeted layer(s):} Data-processing and computation layer (\GC{D1}, \GC{D5}), where model outputs and confidence scores are generated (refer Section~\ref{sec:SystemArch}).

\subsubsection{Side-channel attacks (AV6)}
\noindent\textbf{Mechanism/strategy:} As shown in Table~\ref{tab:AttacksModel}, side channel attacks (SCA) exploit non functional program behavior, such as memory access, cache activity, execution timing, and power consumption, to extract secret information~\cite{phan2017synthesis}. These attacks can compromise confidentiality by recovering network structures and parameters~\cite{hua2022reverse}, and can indirectly threaten integrity by enabling the modification of model weights. Xiang et al.~\cite{xiang2020open} estimate DNN parameter sparsity via a side-channel attack, identifying DNN structures and deriving pre-trained parameter values in embedded devices. Wang et al.~\cite{wang2023side} simulate the dynamic power trace of mixed-signal RRAM in-memory-computing macros, successfully extracting a complete DNN architecture from isolated memory blocks using only the power trace of each tile, without prior knowledge of the model.

\noindent\textbf{Targeted layer(s):} Data-processing and computation layer (\GC{D5}, \GC{D6}), and the underlying hardware executing model inference (refer Section~\ref{sec:SystemArch}).

\subsection{Attacks targeting inputs}
In Figure~\ref{fig:CAVVisionArch}, CAVVS modules (\YC{P1}, \YC{P2}) can be vulnerable to attacks, such as evasion targeting sensor input and MitM intercepting and altering genuine communication~\cite{gupta2023investigation}.

\begin{table}[!ht]
    \centering
    \hyphenpenalty 10000
    \scriptsize
    \hyphenpenalty=10000
    \caption{Attacks targeting the input surface. \textit{D} = direct impact; \textit{ID} = indirect impact; -- = no identified impact.}\label{tab:AttacksInput}
    \begin{tblr}{
        width=\linewidth,
        colspec={
            |p{.04\linewidth}|p{.15\linewidth}|p{.04\linewidth}| p{.04\linewidth}|p{.04\linewidth}|p{.4\linewidth}|
        },
        row{1}={c},
        row{2}={c},
    }\hline
    & & \SetCell[c=3]{c}{\textit{Security property}} & \\\cline{3-5}
    \# & \textbf{Attacks types} & \textbf{C} & \textbf{I} & \textbf{A} & \textbf{Mechanism/strategy} \\\hline
    AV7 & Evasion & -- & \textit{D} & \textit{ID} & Adversarial patches, physical perturbations, spoofed objects, laser interference, and replay\\\hline
    AV8 & Man-in-the-middle & \textit{D} & \textit{D} & -- & Eavesdropping, message interception, replay, and modification\\\hline
    \end{tblr}
\end{table}

\subsubsection{Evasion (AV7)}
\noindent\textbf{Mechanism/strategy:} Evasion attacks subvert model predictions during inference. As shown in Table~\ref{tab:AttacksModel}, these attacks highly impact integrity by corrupting inputs and causing incorrect decisions, and can disrupt availability by overloading CAVVS into operational disruptions or denial-of-service scenarios~\cite{gupta2024visual}. Also referred to as decision-time attacks, they are executed by adversaries who subtly modify inputs to deceive models~\cite{oliynyk2023know}. Physical adversarial object-evasion attacks, in which malicious patches or altered traffic signs obscure or cause misclassification of critical road objects, are a significant threat to CAV perception systems, as detailed through the case studies in Section~\ref{sec:CAVVSChallenges}. Chernikova et al.~\cite{chernikova2019self} demonstrate evasion attacks against neural networks used for steering-angle prediction, using the 2014 Udacity challenge dataset. Spoofing and replay is used to evade LiDAR by injecting signals into target receivers~\cite{wang2024data}; Mahima et al.~\cite{mahima2024toward} discuss laser spoofing via infrared lasers directed at a targeted LiDAR system.

Studies~\cite{deng2021deep} have reported that evasion attacks in both white-box and black-box settings that disturb end-to-end driving, replacing traffic signs or billboards with adversarial versions, or drawing black strips or patterns to impair detection and recognition. White-box methods primarily use gradient-based, optimization-based, and generative model-based techniques that exploit the target model's architecture and parameters. Black-box methods rely on repeated querying to infer model behavior: transfer-based methods train adversarial examples on a surrogate model and transfer them to the target; score-based methods query the target repeatedly to estimate confidence scores; and decision-based methods manipulate input data to alter the model's final decision~\cite{niu2020decade}.

Defending against evasion typically relies on adversarial training or input-sanitization pipelines, both of which introduce a robustness-performance trade-off: hardening a model against perturbed inputs generally reduces clean-data accuracy and increases inference latency, which is a critical constraint for CAVVS given its real-time operational requirements~\cite{deng2021deep}. This trade-off means that the choice of defense cannot be made independently of the CAVVS's latency and accuracy budgets established in Section~\ref{sec:SystemArch}.

\noindent\textbf{Targeted layer(s):} Perception layer (\YC{P1}, \YC{P2}), where sensor inputs are acquired (refer Section~\ref{sec:SystemArch}).

\subsubsection{Man-in-the-middle (AV8)}
\noindent\textbf{Mechanism/strategy:} As shown in Table~\ref{tab:AttacksModel}, MitM attacks compromise confidentiality and integrity by intercepting and altering communication between the perception and data-processing layers~\cite{gupta2023investigation}. Eavesdropping, replay, and data manipulation are the common mechanisms: adversaries can exploit ML-based generative models in CAVs by launching MitM attacks between the data-sensing and classification stages, potentially compromising decision-making~\cite{wang2020man}.

\noindent\textbf{Targeted layer(s):} Interfaces between the perception layer (\YC{P1}, \YC{P2}) and the data-processing and computation layer (refer Section~\ref{sec:SystemArch}).

\subsection{Practical exposure and adversary capability}
Across the eight attack vectors, three practical patterns emerge that are not visible from any single attack in isolation. First, adversary capability and access type form a clear hierarchy. Evasion (AV7) and many MitM variants (AV8) can require only low-to-medium capability, such as physical line-of-sight access, spoofed signals, or access to exposed communication interfaces. In contrast, data poisoning (AV1), logic corruption (AV4), and some side-channel attacks (AV6) generally require privileged access to a data or model pipeline, or sustained physical proximity to the target hardware. Model extraction (AV3) and membership inference (AV5) primarily rely on repeated remote queries and therefore may be easier to scale, although their effectiveness depends on the information exposed by the model interface.

Second, attacker effort and stealthiness are closely tied to the type of access required as described in Table~\ref{tab:SensorComparison}. Physical input attacks can be conducted without a persistent foothold inside the vehicle and can be replicated across vehicles or locations using low-cost artefacts. Model-directed attacks are known to be more difficult to execute initially, but a successfully extracted model or compromised model component can be reused across multiple vehicles or deployments. These distinctions show why practical exposure cannot be assessed solely from the technical sophistication of an attack.

Third, comparing the four model-directed attacks by their consequences rather than by their mechanisms shows that logic corruption poses the most direct safety risk to CAVVS. It can manipulate decision outputs during inference and can be triggered on demand through a backdoor or other compromised processing logic. Model extraction and membership inference primarily threaten confidentiality, including model intellectual property and training-data privacy, without directly altering CAVVS behaviour. Side channel attacks are comparatively less common in practice, but they can act as enabling precursors because an extracted architecture or parameter set can reduce the cost of subsequently crafting logic corruption or evasion attacks against the same model.

From a practical exploitation perspective, the input surface presents the greatest exposure within CAVVS. Unlike many data- and model-directed attacks, which require privileged, remote, or sustained physical access, evasion attacks against sensor inputs can be carried out using low-cost, physically deployable artefacts such as patches, printed signs, or light sources, without access to the underlying model or training pipeline. This disparity between adversary effort and potential impact motivates the practical relevance and benchmarking strategies discussed next.
}

\revisedSG{
\section{Open challenges and research directions}\label{sec:OCRD}
Atomic road events are spontaneous, discrete incidents such as an emergency vehicle approach, pedestrian jaywalking, a sudden lane change, or an abrupt weather shift that stress the CAVVS reference architecture across every layer. Because these events demand immediate perception and response, they amplify the practical impact of input surface attacks, particularly evasion attacks, which have already been identified as requiring the lowest effort while producing the highest consequences. The discussion below therefore relates these events, in general, to the perception layer and input attack surface of the CAVVS architecture, while also highlighting the residual robustness challenges that remain unresolved.

\subsection{Atomic road events}
We outline several atomic road events that present challenges to safe autonomous driving. CAVs encounter these events because of sensor limitations, obstructed line-of-sight, vulnerabilities in deep-learning models, and the lack of up-to-date high-definition maps; each factor maps directly onto the perception layer and the input attack surface of the CAVVS architecture. We group the events into object-detection, normal/spontaneous, and new/unexpected categories.

\subsubsection{Object detection}
\begin{itemize}[leftmargin=*]
    \item \textit{Pedestrian Crosswalk Activation}: Detecting when pedestrians activate a crosswalk signal, indicating their intention to cross the road and requiring the attention of autonomous vehicles.
    \item \textit{Pedestrian Jaywalking}: Detecting cases where pedestrians cross the road outside designated crosswalks or against traffic signals, requiring cautious navigation and potential intervention.
    \item \textit{Emergency Vehicle Presence}: Identifying the presence of emergency vehicles, such as ambulances or fire trucks, with their corresponding lights and sirens activated, necessitating appropriate response and yielding.
\end{itemize}

\subsubsection{Normal and spontaneous situations}
\begin{itemize}[leftmargin=*]
    \item \textit{Lane Change}: Detecting instances where a vehicle changes its lane, either to the left or right, indicating a potential change in trajectory and behavior.
    \item \textit{Lane Departure}: Detecting cases where a vehicle unintentionally or abruptly leaves its designated lane, potentially indicating driver distraction or impairment.
    \item \textit{Traffic Congestion}: Identifying instances of heavy traffic congestion, which can affect the speed and flow of vehicles and require appropriate navigation and planning strategies.
    \item \textit{Vehicle Turn Signals}: Recognizing instances where a vehicle activates its turn signals, indicating an intention to turn left or right, and providing valuable information for predicting its future trajectory.
    \item \textit{Weather Conditions}: Adverse weather conditions, such as snowstorms, sandstorms, or similar phenomena, can result in reduced visibility. 
\end{itemize}

\subsubsection{New and Unexpected situations}
\begin{itemize}[leftmargin=*]
    \item \textit{Road Construction Zones}: Detecting areas of road construction or maintenance, which may require drivers to adjust their speed, change lanes, or follow temporary traffic rules.
    \item \textit{Traffic Light Violation}: Identifying cases where a vehicle runs a red light or fails to stop at a stop sign, which can have significant safety implications.
    \item \textit{Collision Avoidance Maneuvers}: Recognizing situations where a vehicle performs evasive maneuvers, such as sudden braking or swerving, to avoid a collision with another object or vehicle.
    \item \textit{Calamities and Natural Disasters}: Early detection of situations such as earthquakes, floods, lightning strikes, and hailstorms is crucial to avoid potential issues.
\end{itemize}

\begin{figure*}[!ht]
    \captionsetup{format=hang,font=small, margin=3pt}
    \hyphenpenalty 10000
    \centering
    \subfloat[Vehicle is identified as a racer with 12.03\% confidence.]{
        \includegraphics[width=0.24\linewidth]{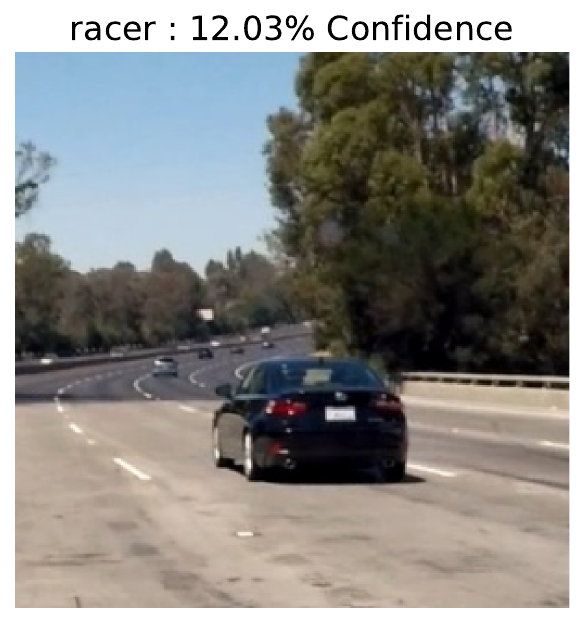}
    }
    \subfloat[Darker surroundings improves model confidence from 12.03\% to 72.57\%.]{
        \includegraphics[width=0.24\linewidth]{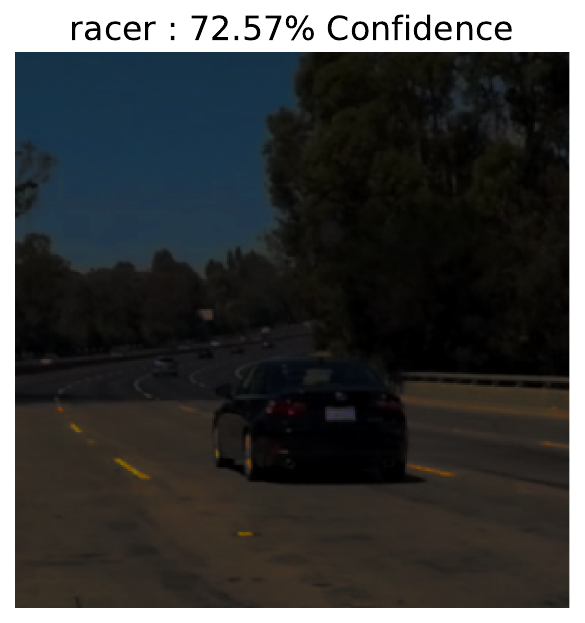}
    }
    \subfloat[Bright light confuses the model, causing the vehicle labeled as a sports car instead of racer.]{
        \includegraphics[width=0.24\linewidth]{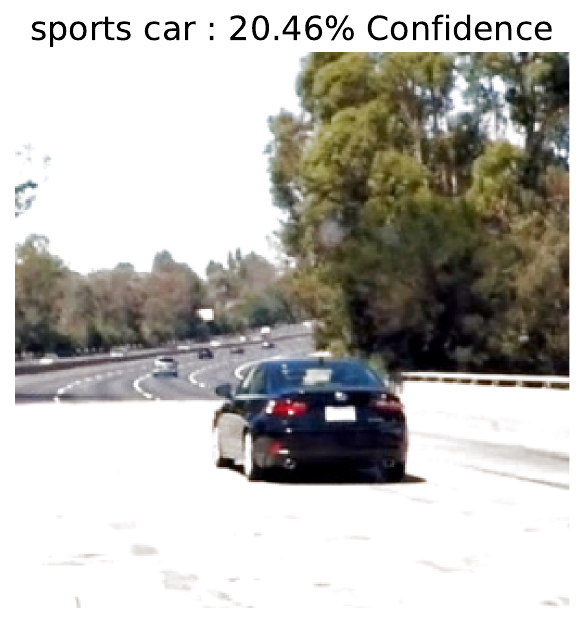}
    }
    \subfloat[Sun flare cause a vehicle to be misidentified as a planetarium.]{
        \includegraphics[width=0.24\linewidth]{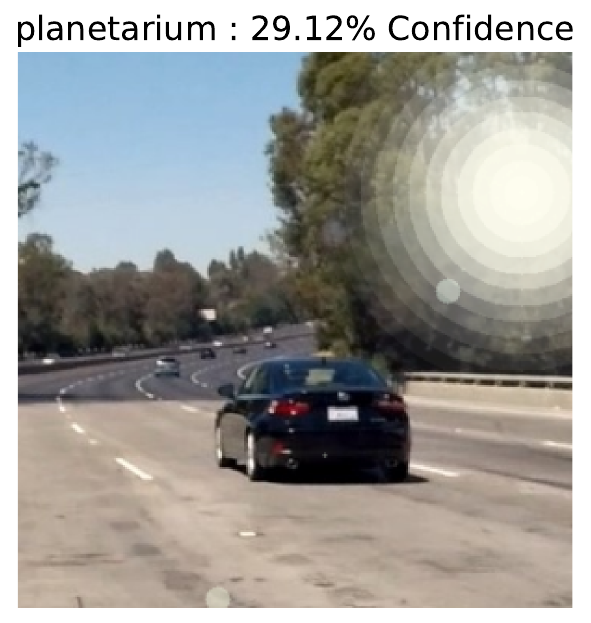}
    }
    \\ 
    \subfloat[Presence of Snow causes the whole scene is detected as geyser.]{
        \includegraphics[width=0.24\linewidth]{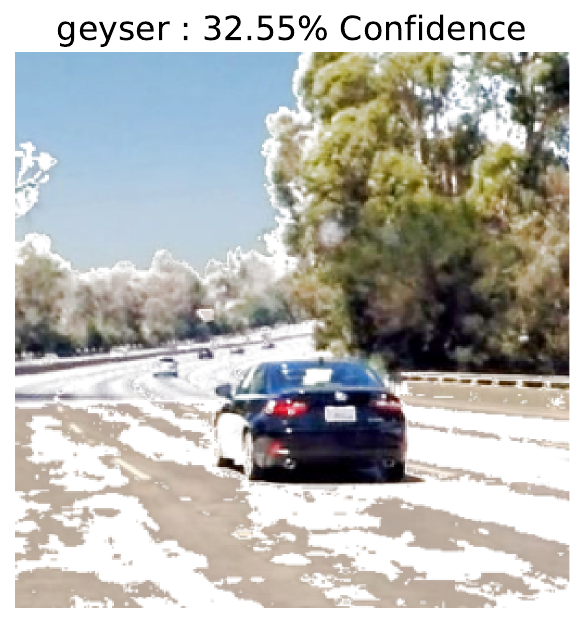}
    }
    \subfloat[A rainy condition is detected as worm fence.]{
        \includegraphics[width=0.24\linewidth]{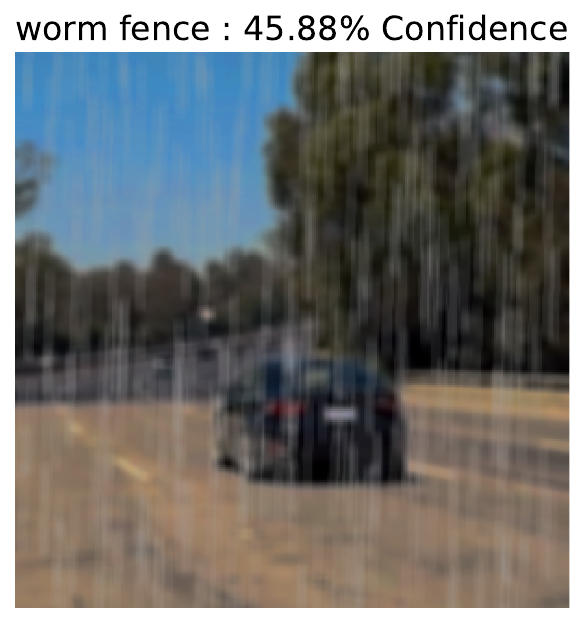}
    }
    \subfloat[In foggy condition, vehicle is detected as trailer truck.]{
        \includegraphics[width=0.24\linewidth]{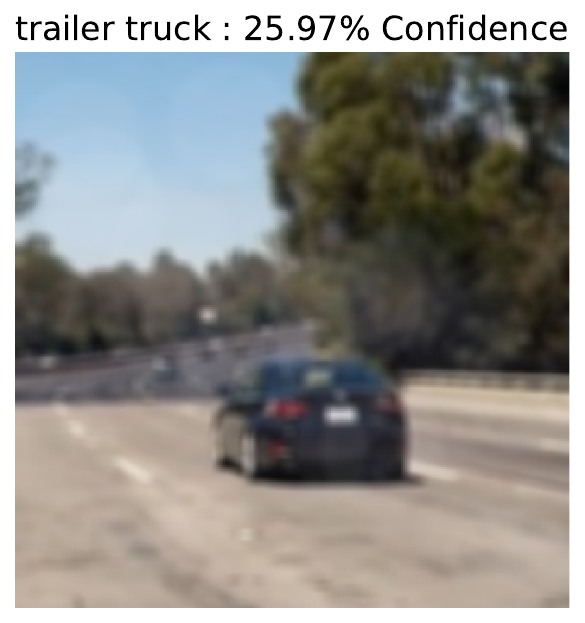}
    } 
    \subfloat[Autumn conditions do not affect the detection capability.]{
        \includegraphics[width=0.24\linewidth]{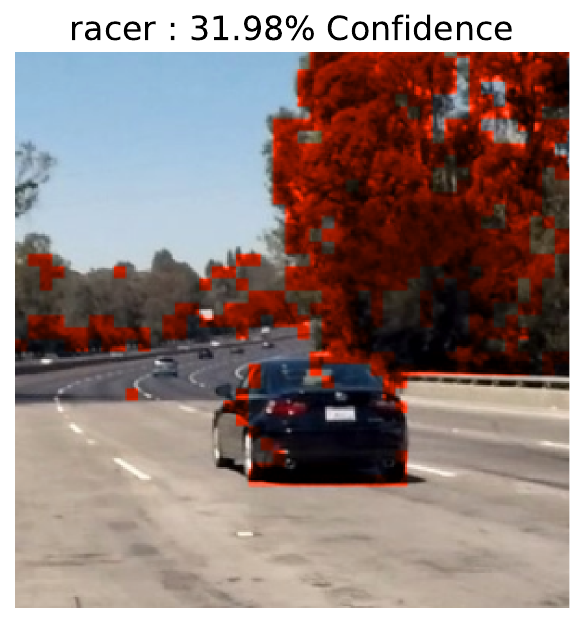}
    }\\
    \subfloat[Random shadows cause a vehicle to be misidentified as a warplane.]{
        \includegraphics[width=0.24\linewidth]{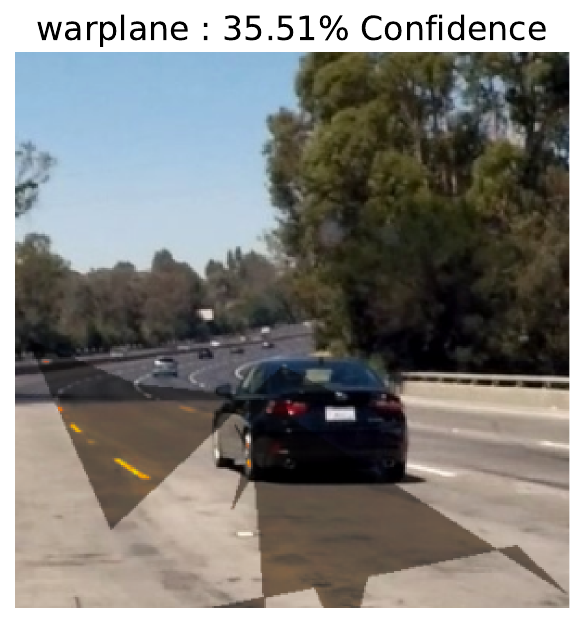}
    }
    \subfloat[Presence of gravel is detected as car mirror.]{
        \includegraphics[width=0.24\linewidth]{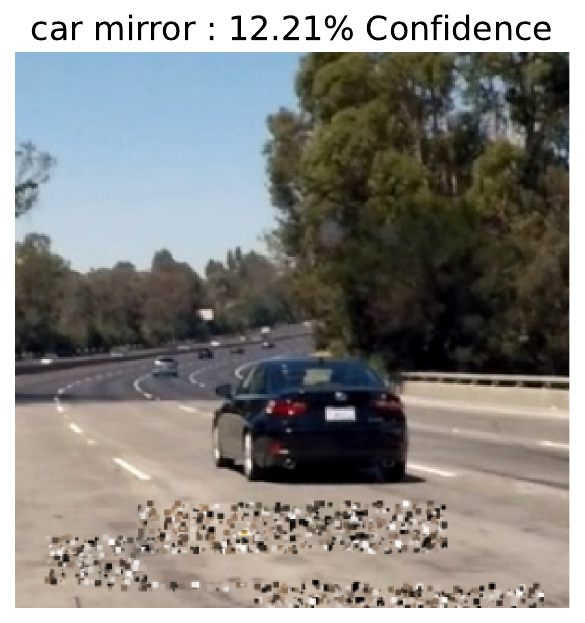}
    }
    \subfloat[A random patch causes vehicle to be identified as a tow truck.]{
        \includegraphics[width=0.24\linewidth]{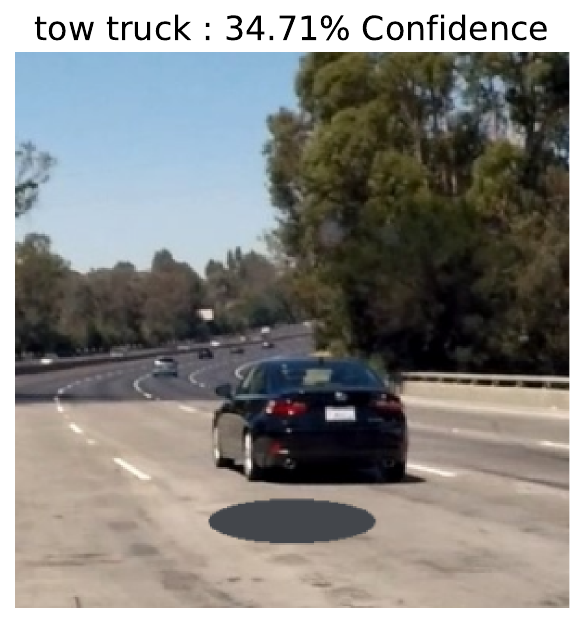}
    }
    \subfloat[Bad road conditions could impact the model categorization accuracy.]{
        \includegraphics[width=0.24\linewidth]{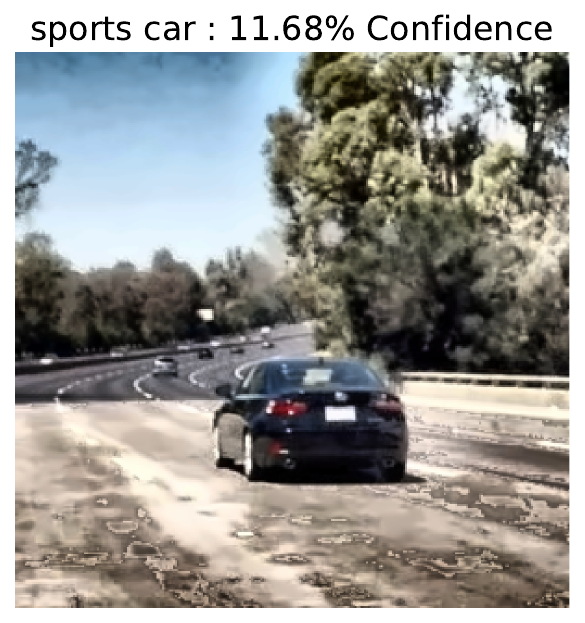}
    }
    \caption{Illustration of the prediction results of the Inception-V3 model~\cite{pytorch2025inception} for an arbitrary input under simulated conditions. These examples are included solely to illustrate the general sensitivity of image classifiers to environmental changes. They do not constitute empirical validation of CAVVS security, as the model was not trained for autonomous-driving perception.}\label{fig:AtomicRoadEvents}
\end{figure*}

\subsubsection{Illustrative environmental sensitivity}
The atomic road events catalogued above highlight a persistent open challenge for CAVVS. Even mature vision models remain highly sensitive to changes in lighting, weather, and scene conditions that commonly accompany these events. To illustrate the nature of this challenge, rather than to claim a formal security evaluation of CAVVS, we present qualitative predictions from a pre-trained Inception-V3 model~\cite{pytorch2025inception} on an arbitrary input subjected to simulated environmental perturbations, as shown in Figure~\ref{fig:AtomicRoadEvents}.

The model initially classifies the vehicle as a racer with only 12.03\% confidence; under darker surroundings the confidence rises to 72.57\%, yet brighter illumination shifts the label to sports car and sun flares produce a planetarium misclassification. Simulated snow, rain, and fog likewise induce gross errors (geyser, worm fence, trailer truck). Random shadows, extraneous patches, and poor road surfaces further degrade accuracy. These uncontrolled examples are not an adversarial benchmark, nor do they employ an autonomous-driving-specific architecture or dataset; they simply underscore that out-of-distribution conditions typical of atomic road events can still defeat contemporary classifiers. Closing this robustness gap within the real-time and multi-modal constraints of the CAVVS perception layer remains an open research problem.

\subsection{Toward standardized robustness benchmarks}
The sensitivity illustrated in Figure~\ref{fig:AtomicRoadEvents} points to a broader methodological gap: existing studies use heterogeneous datasets, models, threat models, and metrics, which limits direct comparison and practical guidance for system designers. Future work therefore requires a benchmark-oriented evaluation framework for CAVVS robustness that connects atomic road events to standardized datasets, attack scenarios, and safety-relevant metrics.

\paragraph{Datasets and scenarios}
Robustness evaluation can move beyond generic image classification toward datasets and scenarios that reflect autonomous driving. Candidate resources include KITTI, BDD100K, nuScenes, and Waymo Open Dataset for 2D/3D detection and tracking, together with simulation platforms such as CARLA, LGSVL/SVL, and AirSim for controlled environmental perturbations and closed-loop testing. Each atomic road event category (object detection, normal/spontaneous, and new/unexpected) can be mapped to concrete scenarios, for example pedestrian crossing under occlusion, emergency-vehicle approach with siren and flashing lights, lane change in dense traffic, and sudden weather transitions.

\paragraph{Metrics and reporting}
A benchmark for CAVVS robustness is expected to report multiple complementary dimensions:
\begin{itemize}[leftmargin=*]
    \item \textit{Perception accuracy:} mAP, NDS, AMOTA, class-wise precision and recall, and localization error under clean, corrupted, and adversarial conditions.
    \item \textit{Environmental robustness:} performance by weather, lighting, visibility, and severity level, including degradation curves rather than single-point results.
    \item \textit{Adversarial evaluation:} attack success rate, perturbation budget, transferability, query count, and robustness under white-box and black-box settings.
    \item \textit{Safety impact:} collision rate, time-to-collision, emergency-braking rate, lane-departure frequency, and intervention rate in closed-loop simulation.
    \item \textit{System constraints:} inference latency, energy consumption, communication overhead, and compatibility with sensor fusion and fallback modes.
\end{itemize}

\paragraph{Evaluation protocols}
A standardized protocol can specify:
\begin{itemize}[leftmargin=*]
    \item the perception model and training protocol (e.g., 2D detector, 3D detector, multi-sensor fusion, BEV fusion, or end-to-end driving stack);
    \item the environmental and adversarial perturbations applied, including severity levels and random seeds;
    \item the attack model, including adversary knowledge, access type, and perturbation budget;
    \item the baseline defences or redundancy mechanisms under test;
    \item the number of trials and statistical analysis used to estimate uncertainty.
\end{itemize}

Such protocols can enable reproducible comparison across studies and support the development of robustness requirements for CAVVS components.

\paragraph{Integration with CAVVS}
Within the CAVVS reference architecture, a benchmark-oriented evaluation framework can be organized by attack surface and lifecycle stage. For the input surface, benchmarks can focus on sensor-level perturbations, spoofing, and adversarial examples under atomic road events (AV7, AV8). For the data and model surfaces, benchmarks can address poisoning (AV1), logic corruption and domain shift (AV4), model extraction (AV3), and side-channel leakage (AV6), including runtime monitoring under realistic deployment resource constraints.

\section{Conclusions}
CAVVS robustness is crucial for achieving Level-5 autonomy and accelerating the commercialization of CAVs. This paper introduced a CAVVS reference architecture spanning the perception, data-processing, and motion-estimation layers, and used it to derive three attack surfaces: data, models, and inputs. These attack surfaces are derived from an explicit threat model of CAVVS assets, adversary capabilities, and lifecycle phases. In contrast to prior CIA-oriented surveys, which treat compliance or CAV components generically, this framework ties attack surfaces directly to the vision-processing pipeline.

Building on this architecture, we analyzed eight attack vectors, data poisoning, exfiltration, extraction, logic corruption, inference attacks, side-channel attacks, evasion, and man-in-the-middle attacks, each characterized by mechanism, targeted layer, and CIA impact, and further compared across adversary capability and practical exposure. This comparison identifies the input surface as the most exposed in practice and logic corruption as the most direct safety risk among model-directed attacks. We also outline atomic road events that exacerbate CAVVS robustness and, motivated by the limitations of illustrative environmental-sensitivity testing, propose a benchmark-oriented framework, i.e., datasets, metrics, and evaluation protocols, for standardized robustness assessment.

This work offers security practitioners and researchers a structured basis for vulnerability assessment, defense-tradeoff analysis, and future empirical validation of CAVVS.
}


\section*{Acknowledgment}
This work is funded by the UK Government through the New Deal for Northern Ireland. The funding is delivered on behalf of the Northern Ireland Office and the Department for Digital, Culture, Media and Sport by Innovate UK.

\bibliographystyle{ieeetr} 
\bibliography{references}

\vspace{-5mm}

\begin{IEEEbiography}[{\includegraphics[width=1in,height=1.25in,clip,keepaspectratio]{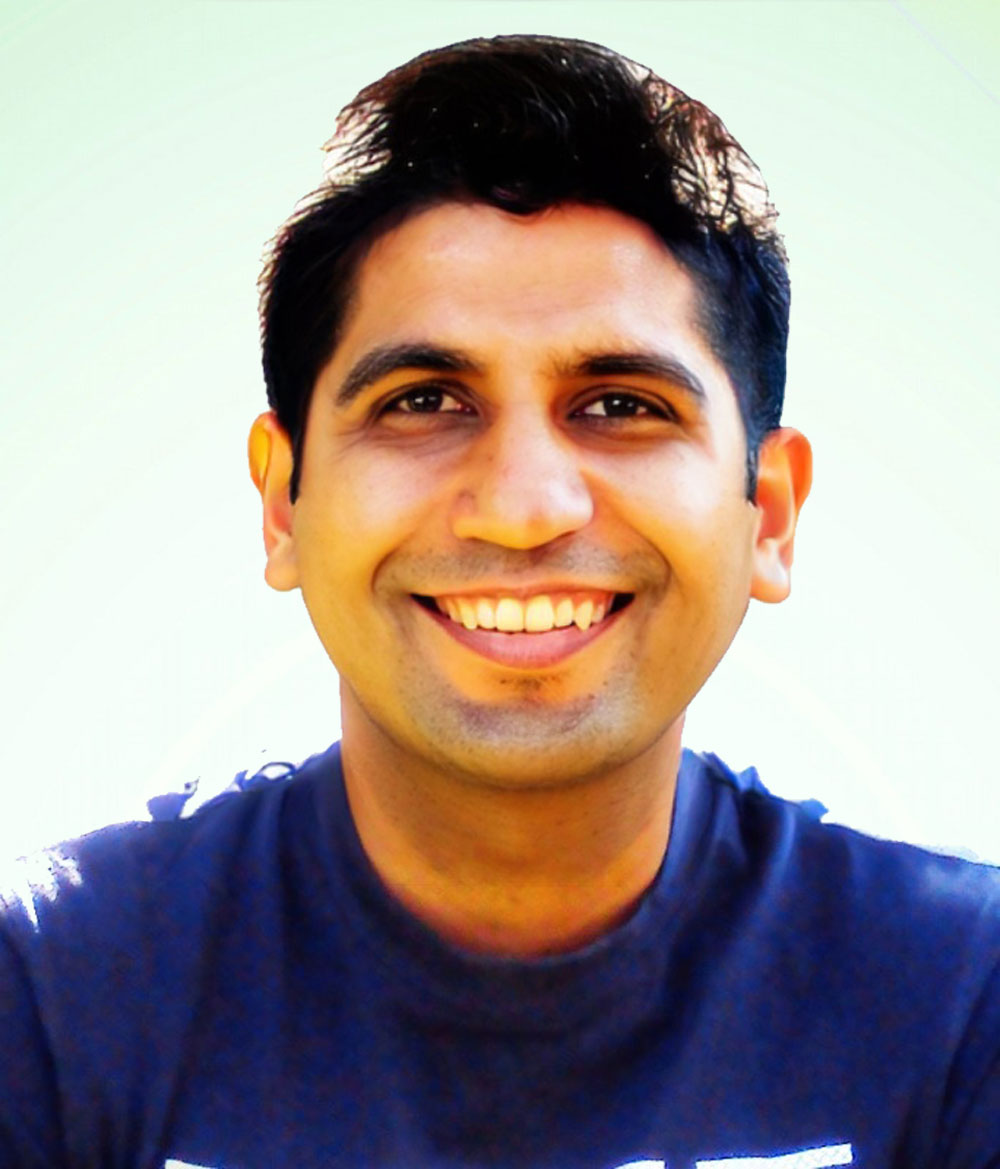}}]{Sandeep Gupta}
received his Ph.D. degree in Information \& Communication Technology from the University of Trento, Italy. He is currently a Principal Engineer with Queen's University Belfast, U.K. From 1999 to 2016, he was with Samsung, Accenture, and Mentor Graphics (now Siemens) in information technology domains driving several research projects and products from their incubation to next-generation iteration. Since 2016, he has worked on UKRI and EU H2020 projects, including CyberSec4Europe, Collabs, E-Corridor, and NeCS. He has published more than 30 papers in peer-reviewed journals and leading conferences. His research interests include secure-by-design AI systems, trustworthy AI, biometric-based identity and access mechanisms, and usable security and privacy solutions for cyber-physical systems and the IoT. He was a recipient of the prestigious Marie Sklodowska-Curie Research Fellowship.
\end{IEEEbiography}

\begin{IEEEbiography}[{\includegraphics[width=1in,height=1.25in,clip,keepaspectratio]{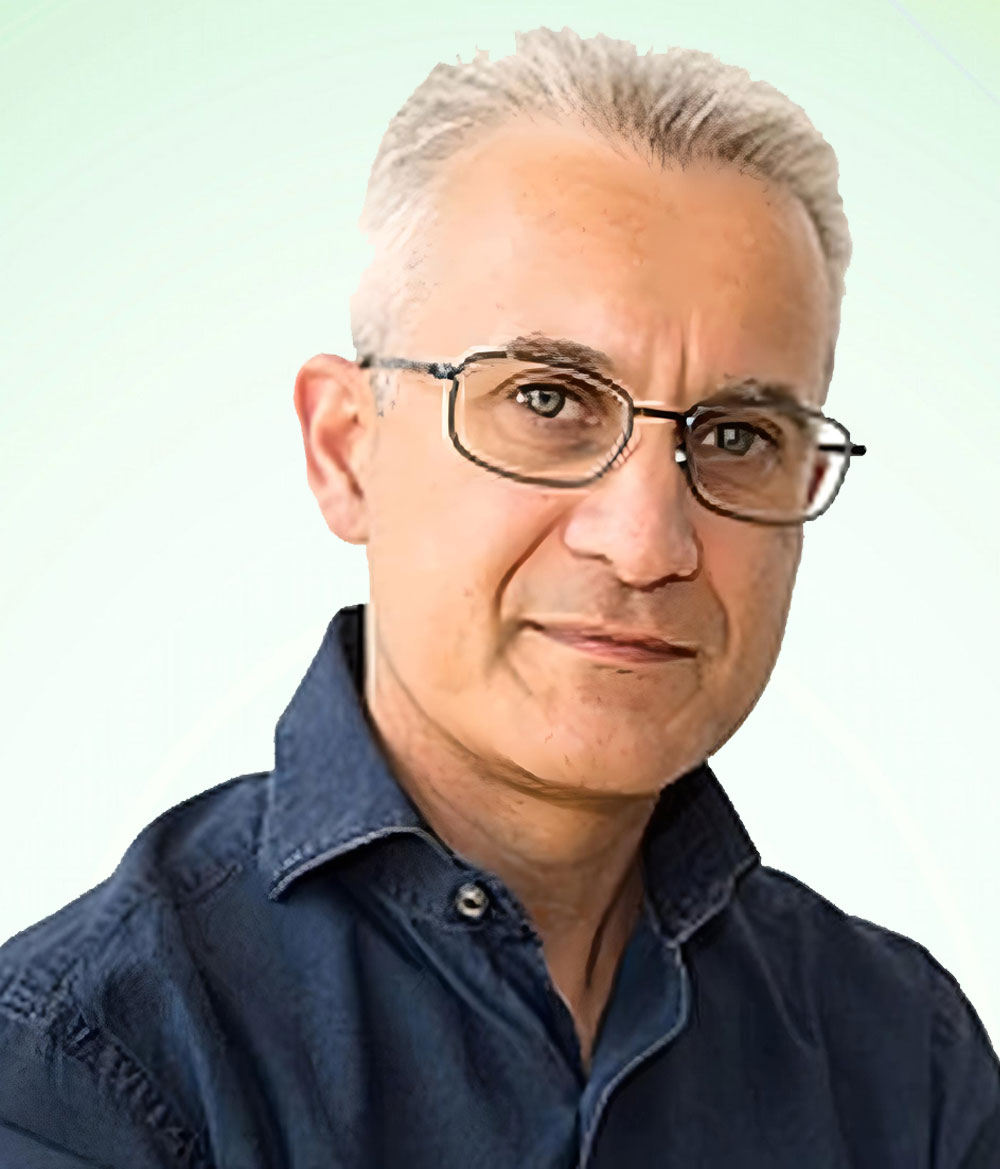}}]{Roberto Passerone} (S'96-M'05) received the M.S.\ and Ph.D.\ degrees in electrical engineering and computer sciences from the University of California, Berkeley, in 1997 and 2004, respectively. He is currently a Professor of electronics with the Department of Information Engineering and Computer Science, University of Trento, Italy. Before joining the University of Trento, he was a Research Scientist with Cadence Design Systems. He has published numerous research articles at international conferences and journals in the area of design methods for systems and integrated circuits, formal models, and design methodologies for embedded systems, with particular attention to image processing, and wireless sensor networks. He has participated in several European projects on design methodologies, including SPEEDS, SPRINT, and DANSE, and he was the Local Coordinator for ArtistDesign, COMBEST, and CyPhERS. He has served as Track Chair for the Design Modeling and Verification for Embedded and Cyber-Physical Systems track at DATE, and was General Chair and Program Chair for various editions of SIES.
\end{IEEEbiography}
\vfill

\end{document}